\documentclass[letterpaper]{article} % DO NOT CHANGE THIS
\usepackage{aaai2026}  % DO NOT CHANGE THIS
\usepackage{times}  % DO NOT CHANGE THIS
\usepackage{helvet}  % DO NOT CHANGE THIS
\usepackage{courier}  % DO NOT CHANGE THIS
\usepackage[hyphens]{url}  % DO NOT CHANGE THIS
\usepackage{graphicx} % DO NOT CHANGE THIS
\usepackage{natbib}  % DO NOT CHANGE THIS AND DO NOT ADD ANY OPTIONS TO IT
\usepackage{caption} % DO NOT CHANGE THIS AND DO NOT ADD ANY OPTIONS TO IT
\usepackage{algorithm}
\usepackage{algorithmic}
\usepackage{amssymb}
\usepackage{amsmath}
\usepackage{booktabs}
\usepackage{multirow}
\usepackage{makecell}
\usepackage{pifont}
\usepackage{array}
\usepackage{dsfont}
\usepackage{float}
\usepackage{color}
\usepackage{colortbl}
\usepackage{xcolor}
\usepackage{epsfig}
\usepackage{subcaption}
\usepackage{newfloat}
\usepackage{listings}
\DeclareCaptionStyle{ruled}{labelfont=normalfont,labelsep=colon,strut=off} % DO NOT CHANGE THIS
\floatstyle{ruled}
\newfloat{listing}{tb}{lst}{}
\floatname{listing}{Listing}
\nocopyright
\title{MaskAnyNet: Rethinking Masked Image Regions as Valuable Information in Supervised Learning}
\author{
    Jingshan Hong\textsuperscript{\rm 1},
    Haigen Hu\textsuperscript{\rm 1}\thanks{Corresponding author},
    Huihuang Zhang\textsuperscript{\rm 1},
    Qianwei Zhou\textsuperscript{\rm 1},
    Li Zhao\textsuperscript{\rm 2},
}
\affiliations{
    \textsuperscript{\rm 1}College of Computer Science and Technology, Zhejiang University of Technology\\
    Hangzhou, Zhejiang, 310014, China\\
    \textsuperscript{\rm 1}Zhejiang Normal University\\
    Jinhua, Zhejiang, 321000, China\\
    2112101525@zjut.edu.cn, hghu@zjut.edu.cn, huihuang@zjut.edu.cn, zqw@zjut.edu.cn, zhaoli2023@zjnu.edu.cn
}

\usepackage{bibentry}
\begin{document}

\maketitle

\begin{abstract}
In supervised learning, traditional image masking faces two key issues: (i) discarded pixels are underutilized, leading to a loss of valuable contextual information; (ii) masking may remove small or critical features, especially in fine-grained tasks.
In contrast, masked image modeling (MIM) has demonstrated that masked regions can be reconstructed from partial input, revealing that even incomplete data can exhibit strong contextual consistency with the original image. This highlights the potential of masked regions as sources of semantic diversity.
Motivated by this, we revisit the image masking approach, proposing to treat masked content as auxiliary knowledge rather than ignored. Based on this, we propose MaskAnyNet, which combines masking with a relearning mechanism to exploit both visible and masked information. It can be easily extended to any model with an additional branch to jointly learn from the recomposed masked region. This approach leverages the semantic diversity of the masked regions to enrich features and preserve fine-grained details. Experiments on CNN and Transformer backbones show consistent gains across multiple benchmarks. Further analysis confirms that the proposed method improves semantic diversity through the reuse of masked content. The demo code is publicly available at: https://github.com/HuHaigen/MaskAnyNet-Code
\end{abstract}


\section{Introduction}
Deep learning has been the cornerstone of progress in visual recognition tasks, driving remarkable advances in image classification, detection, and segmentation through large-scale annotated datasets and more advanced model architectures. \cite{I1,I2,I3,I4}. Image masking methods, as a common data augmentation approach, have been widely adopted in supervised learning and self-supervised image reconstruction \cite{I5,I6,I7}. These methods improve model robustness by occluding parts of the image, forcing the model to make predictions or reconstructions based on incomplete information.
\begin{figure}[t]
\centering %表示居中
\includegraphics[width=8.5cm,height=6cm]{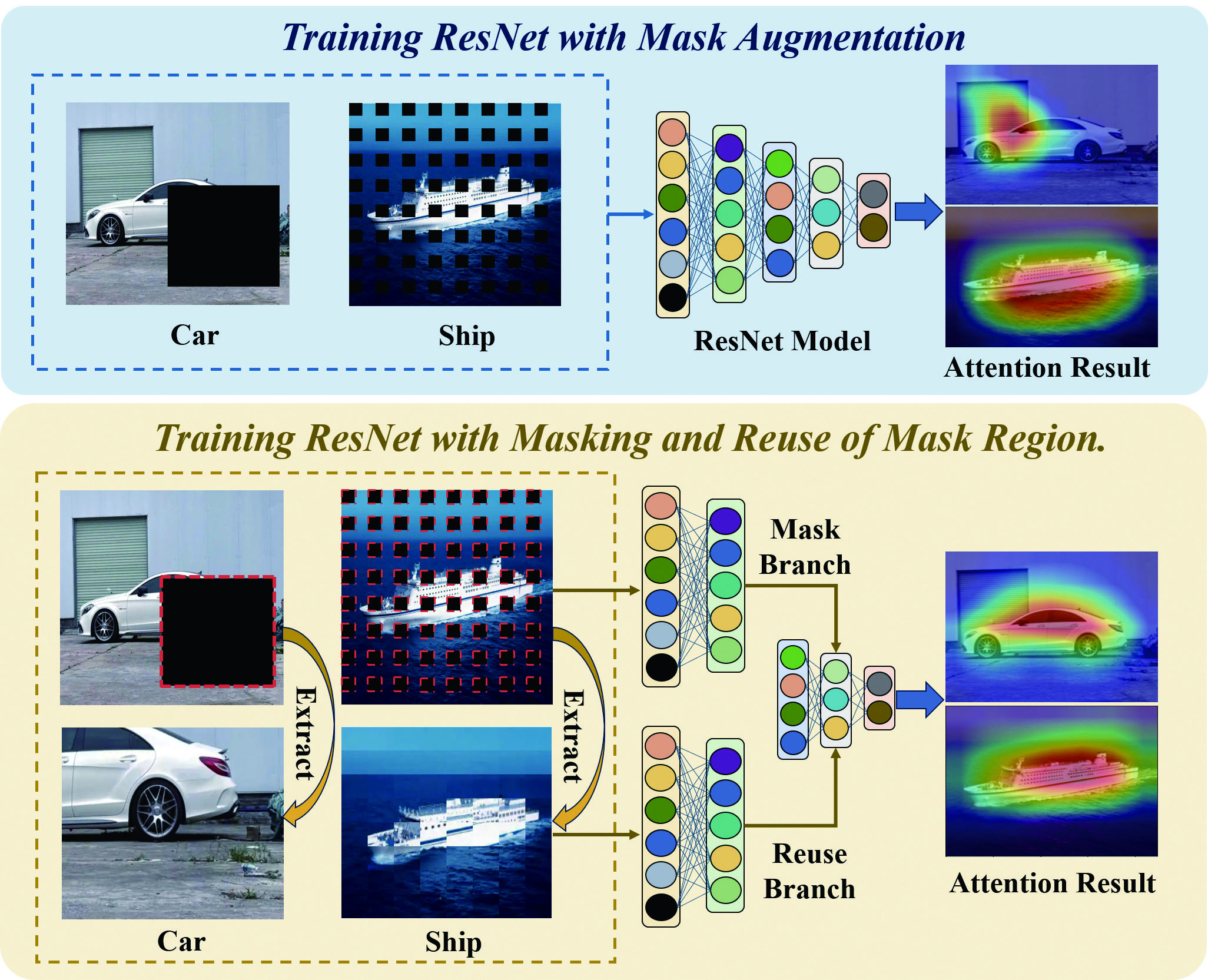}
\caption{Comparison of effect between conventional mask discarding and with reuse strategies for complementary visual information based on ResNet-34. Top: the heatmap can only cover part of the target and fails to clearly perceive edge details of the target. (e.g., partial coverage of vehicles or distraction by background elements like ocean scenes). Bottom: the heatmaps can precisely focus on the target area by repurposing these masking regions as complementary visual information sources.} 
%Comparison of traditional ResNet and ResNet with masking and masked region reuse. Compared to ResNet, our approach adopts a dual-branch architecture that leverages masked information to improve target perception.}
\label{Fig1}
\end{figure}
Although these methods have shown effectiveness in regularization and self-supervised representation learning, in the field of supervised learning, they usually treat masked pixel information as useless noise rather than a reconfigurable and reusable source of information. This will lead to two critical limitations: (i) Direct discarding of masked region pixel fails to fully exploit available image information, thereby constraining the model's input utilization efficiency; (ii) Critical semantic elements (e.g. small objects and fine-grained textures) may be obscured, which inhibits feature learning in such regions and consequently compromises recognition capability for small or occluded targets.

%The information within the masked regions is directly discarded during training, resulting in insufficient exploitation of the available image data and limiting the model's efficiency in utilizing input information; (ii) semantically critical regions, such as small objects, boundaries, or fine details may be occluded; this will limit the model's ability to learn the features of these regions, which subsequently impairs its capacity to recognize small or occluded targets.
We note that masked image modeling (MIM) enables the reconstruction of comprehensive visual representations from minimal visible pixel subsets \cite{I8,I9,I10}, which means that even a limited set of unmasked patches can preserve the global structural semantics of the original image \cite{I11}.
%To address these limitations, we draw inspiration from masked image modeling (MIM), which enables the reconstruction of comprehensive visual representations from minimal visible pixel subsets \cite{I8,I9,I10}. These findings suggest that even a limited set of unmasked patches can preserve the global structural semantics of the original image \cite{I11}.
%Motivated by these observations, we reconsider masked image regions in supervised learning: instead of treating them as discarded noise, we argue that masked regions can be reused as a complementary source of visual information. Can we utilize image masking as a mechanism to provide diverse and fine-grained features, thereby enhancing feature representation learning?
Motivated by these observations, we rethink masked image regions in supervised learning: rather than discarding them as noise, we propose to repurpose these regions as complementary visual information sources. Specifically, we explore whether image masking can serve as a mechanism to provide diverse and fine-grained features, thus enhancing feature representation learning.

%we leverage image masking as a mechanism for providing diverse and fine-grained features that enhance feature representation learning.

% We argue that masked regions can be reused as a complementary source of diverse visual information, potentially improving generalization and robustness.

% Motivated by these observations, we adopt a different perspective: rather than employing masking merely to discard information, we leverage image masking as a mechanism for providing diverse and fine-grained features.
Consequently, we conducted preliminary investigations along this direction. Figure \ref{Fig1} illustrates the comparative effects of reusing masked regions as complementary visual information sources by using class activation heatmaps. From Figure \ref{Fig1}, the traditional masking approach fails to capture complete features and perceive the clear details of the edge, thus limiting the ability of the model to make accurate predictions. While the reuse strategy for complementary visual information can achieve precise target localization with full-contour awareness while effectively suppressing irrelevant background noise, it means that the information reuse auxiliary branch recovers these details through masked latent learning.

Based on these insights, we propose a dual-branch architecture, called MaskAnyNet, to repurpose discarded mask regions as complementary visual information sources. Specifically, we redesign backbone models into a dual-branch architecture that simultaneously learns from both visible content and masked regions. A primary branch processes the masked image to extract global features, while another auxiliary branch extracts fine-grained details from masked regions to refine feature representations. This integrated design leverages both the regularization benefits of masking and enhanced representational capacity from reintegrated local features, enabling comprehensive pixel-level information utilization and multi-perspective visual understanding.
\textbf{The main contribution of this work can be summarized as follows}.
\iffalse
\begin{itemize}
\item We propose MaskAnyNet, a unified architecture with a mask-guided reuse branch, compatible with both CNNs and Transformers. Across various datasets, our method consistently improves baseline performance while maintaining comparable computational efficiency.
\item We systematically analyze our masking and reuse patterns, providing configuration guidelines for our method. We further quantify the effectiveness of different masking strategies using information entropy and similarity metrics, offering theoretical support for our approach.
\end{itemize}
\fi
\begin{itemize}
\item \textbf{Architecture Innovation}: We propose \textbf{MaskAnyNet}, a novel unified architecture featuring a \textit{mask-guided information reuse branch}. Compatible with both CNNs and Transformers, it consistently boosts baseline performance across diverse datasets while maintaining computational efficiency comparable to vanilla backbones.
\item \textbf{Mask Pattern Analysis}: We rigorously characterize masking and reuse patterns, establishing generalized configuration guidelines. Using \textit{information entropy} and \textit{cross-region similarity metrics}, we quantitatively validate mask strategy effectiveness, providing theoretical grounding for our approach.
\end{itemize}

\section{Relate Work}
\subsection{Image Masked Data Augmentation}
Image masking is a widely used method in the field of computer vision\cite{r1, r2, R22, R23}. By masking specific areas of the input image, the model focuses more attention on the area of interest, thereby improving the generalization ability of the model. In the field of data augmentation, some typical methods such as CutOut\cite{r3} and Random Erasing\cite{r4}, randomly mask parts of the image to encourage the model to focus on different parts of the target, preventing over-reliance on certain local features and significantly improving the model's generalization ability. Building on this, Hide-and-Seek\cite{r5} employs a broader range of random occlusion to enable the model to learn features from more areas of the target, thus improving its recognition ability when the target is occluded or incomplete. Furthermore, GridMask\cite{r6}adopts a structured occlusion strategy by regularly generating grid-shaped occlusions, achieving a better balance between information removal and retention. FenceMask\cite{r7} generates continuous occlusion areas in a fence-like shape, making it more suitable for processing scenes with small targets. It is highly effective in preserving small object features and can be better adapted to datasets with complex small target features.

\begin{figure*}[ht]
\centering %表示居中
\includegraphics[width=16cm,height=8cm]{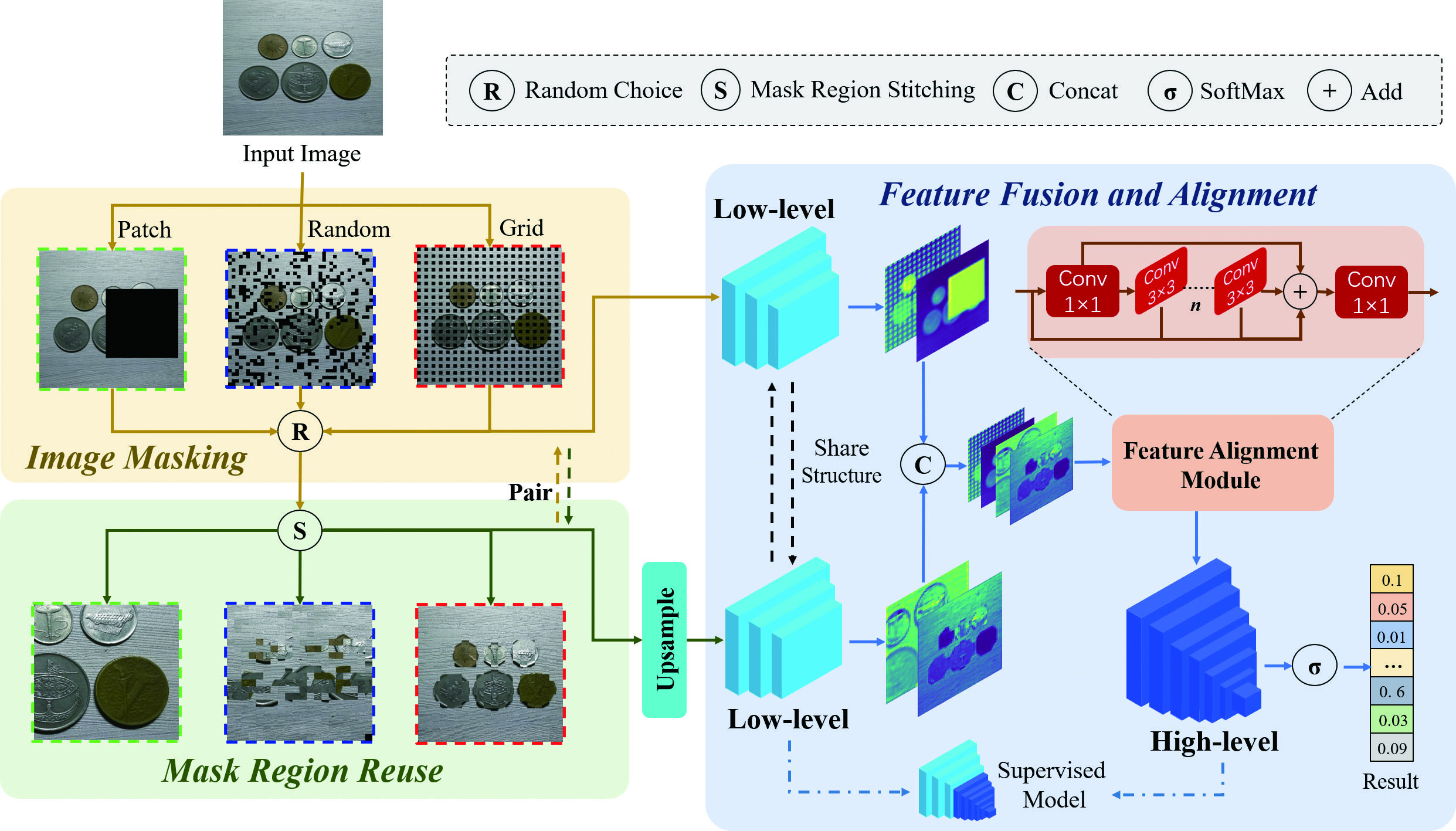}
\caption{Overall architecture of the proposed MaskAnyNet, which consists of three main components: mask generation, mask-region information reuse, and feature fusion and alignment.}
\label{Fig2}
\vspace{-10pt}
\end{figure*}

\subsection{Masked Image Modeling}
%MIM has become a cornerstone of self-supervised learning, widely used in image restoration\cite{R18,R19,R20}, and learning transferable visual representations\cite{r17,R21}. Inspired by Masked Language Modeling (MLM) in NLP, MIM learns visual features by reconstructing occluded image regions without annotations\cite{r8}. BEIT\cite{r9} follows the idea of BERT\cite{r10}, randomly masking patches in the image to predict masked patch representations, significantly improving feature quality. BEITv2\cite{r11} further enhances performance through joint optimization and refined semantic modeling.
MIM has emerged as a cornerstone of self-supervised learning, enabling both image restoration \cite{R18,R19,R20} and transferable visual representation learning \cite{r17,R21}. Inspired by NLP's Masked Language Modeling (MLM), MIM reconstructs occluded regions to learn annotation-free visual features \cite{r8}. BEIT \cite{r9} extends BERT's approach\cite{r10} by predicting representations of randomly masked patches, significantly enhancing feature quality. BEITv2 \cite{r11} further advances performance through joint optimization and semantic refinement.
%MAE\cite{r12} forces the model to reconstruct the complete image from the remaining information. Thereby enabling it to deeply learn the feature representations of the image. SimMIM\cite{r13} allows the encoder to simultaneously perform representation learning and prediction of the masked area, making the model lighter. Furthermore, OmniMAE\cite{r14} extends MAE to the multimodal domain, supporting joint modeling of both video and images. PixMIM\cite{r15} proposes directly reconstructing the pixel values of the occluded areas as the training target, eliminating the dependence on pre-trained visual tokenizers. MixMask\cite{r16} introduces a dynamic distance loss based on CutMix, which calculates similarity based on the mask ratio of each pair of input images, enabling a more accurate measurement of image similarity.
MAE\cite{r12} reconstructs full images from partial inputs to learn deep representations. SimMIM\cite{r13} unifies masked region prediction and representation learning in a single encoder, reducing model complexity. OmniMAE\cite{r14} extends MAE to multimodal video-image joint modeling. PixMIM\cite{r15} directly reconstructs occluded pixel values, eliminating dependency on pretrained tokenizers. MixMask\cite{r16} introduces a CutMix-based dynamic distance loss that computes similarity through mask ratios for precise image matching.
% ConvNextV2\cite{r17} inspired by MAE and uses masked convolution-based encoding and decoding to better capture key features, improving adaptability and robustness to image variations. 
% These methods demonstrate the powerful potential of the masking approach and offer new possibilities for designing visual models in other fields.

% These methods highlight the success of masking in data augmentation and self-supervised learning. However, its exploration in supervised model architectural design remains limited. In supervised learning, masking is typically used for data augmentation, yet masked regions are usually underutilized and risk discarding critical details. This motivates us to rethink MIM beyond beyond self-supervised settings.

%Although masking has achieved great success in data augmentation and self-supervised learning, its potential in supervised settings remains limited, where masking serves only as an augmentation method, leading to the loss of valuable information. This motivates us to rethink how to fully leverage masked signals within supervised learning.
While masking excels in data augmentation and self-supervised learning, its potential in supervised settings remains limited by treating masked regions as disposable noise, causing valuable information loss. This motivates our fundamental reexamination of masked signal utilization within supervised paradigms.

\section{Methodology}
We propose MaskAnyNet, a novel architectural paradigm that integrates region masking with a reuse branch to reintroduce masked pixel information into the learning process.
As shown in Figure \ref{Fig2}, the framework comprises three components: image masking, mask region reuse, and feature fusion and alignment. First, the image masking branch occludes parts of the input using various masking strategies. Then, the reuse branch uses the masked regions to enhance contextual understanding. Finally, the fusion and alignment module integrates features from both branches to optimize their interaction. The details of each component are described below.

\begin{figure}[htbp]
    \centering
    \begin{subfigure}[b]{0.16\textwidth}
        \includegraphics[height=1.6cm,width=2.75cm]{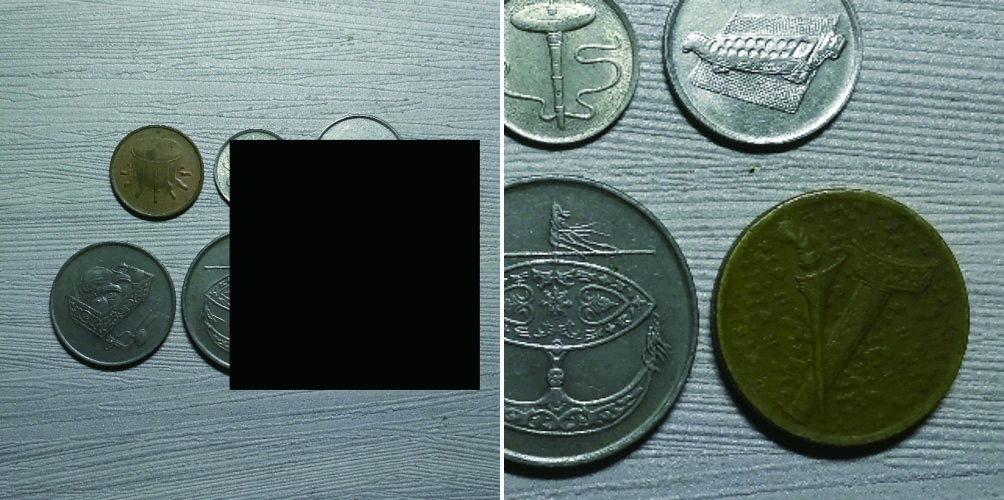}
        \caption{Patch Mask}
        \label{Patch}
    \end{subfigure}%
        \begin{subfigure}[b]{0.16\textwidth}
        \includegraphics[height=1.6cm,width=2.75cm]{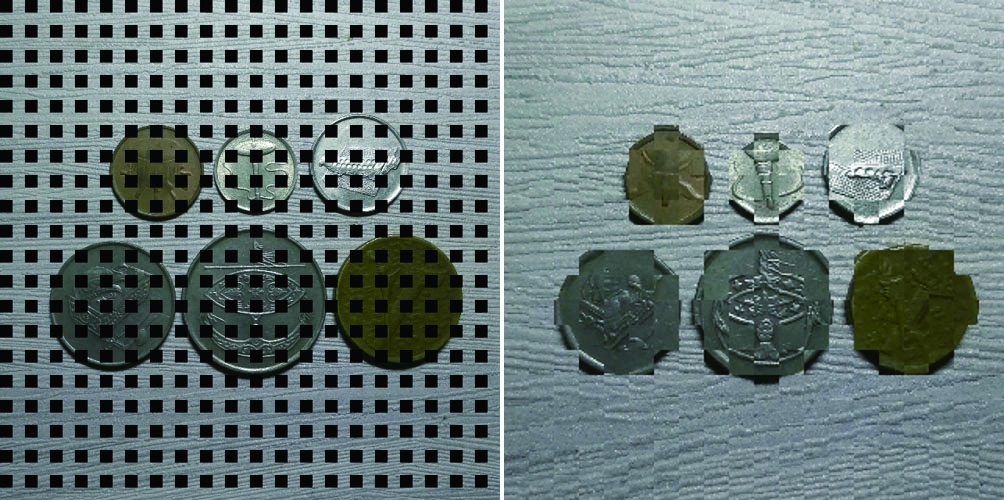}
        \caption{Grid Mask}
        \label{Grid}
    \end{subfigure}%
        \begin{subfigure}[b]{0.16\textwidth}
        \includegraphics[height=1.6cm,width=2.75cm]{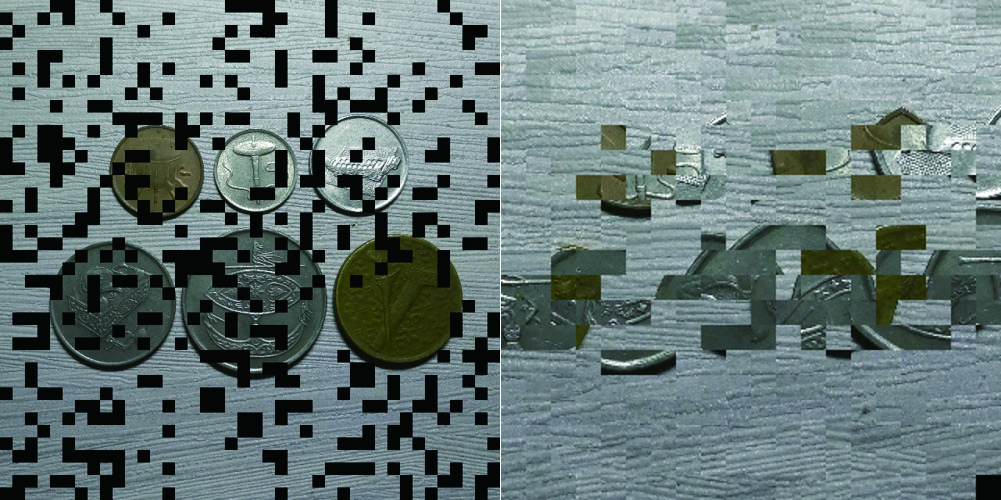}
        \caption{Random Mask}
        \label{Random}
    \end{subfigure}%
    \caption{Comparison of different masking methods. Patch Masking preserves local details, Grid Masking preserves global semantics through fixed-pattern, and Random Masking enhances diversity through irregular patterns.}
    \label{Fig3}
\end{figure}

\subsection{Image Masking Branch}
\label{Image Masking}
The image masking methods can be classified into three types: single Patch mask, Random mask, and grid mask, as illustrated in Figure \ref{Fig3}.
As shown in the figure, grid masking can be regarded as a uniform sampling strategy. When the masked regions are stitched together, it helps preserve essential global structure. This allows the model to efficiently model multi-scale features and enhance the effective receptive field from limited visible pixels.The regular spatial coverage of grid masking also facilitates the extraction of long-range dependencies with minimal information loss.
In contrast, the patch masking strategy emphasizes local detail. By randomly masking fine-grained patches, the model can learn diverse local features such as textures, edges, and small-scale object parts. This improves the model’s sensitivity to fine details and boosts its ability to represent small or irregular targets.
Random masking introduces high variability in the spatial distribution of masked regions, which  encourages the model to reason over more complex and unpredictable patterns. However, due to its unstructured nature, random masking introduces significant spatial distortion, such as positional ambiguity and geometry shifts. Although this property benefits self-supervised tasks like masked image reconstruction, it may undermine performance in supervised tasks like classification or detection, where consistent semantic coherence and spatial alignment are crucial.
Based on the above analysis, our masking strategy integrates both grid and random patch masking. By combining grid and patch masking, it strikes a balance between global semantic preservation and local detail enhancement, enhancing the ability of model to learn diverse visual patterns.

\subsection{Mask Region Reuse Branch}
Although the mask image can serve as a regularization, it can result in the loss of key features, particularly in small target areas. Furthermore, the effectiveness of masking strongly depends on the mask strategy and mask ratio, which may not be universally suitable for all tasks or image distributions. To address these limitations, we design a mask information reuse branch, which is designed to re-learn and enhance the information lost due to masking.
Specifically, let $M$ be the set of masked regions in the input image, where each masked block is denoted as $M_i$, with $i \in[1,2,...,N]$, where $N$ is the total number of masked blocks. The masked blocks are indexed based on their position in the original image. After that, identify all the masked blocks $M_i$ within the input image and extract their corresponding unmasked regions $Ui$ from the original image that spatially corresponds to the masked block $M_i$.

After extracting the masked regions, we combine all extracted patches $U_i$ according to their relative spatial positions in the original image. The reuse image $R$ is then constructed by Eq.\ref{R}:
\begin{equation}
R = \mathcal{S} \left( \{ (U_i, P_i) \}_{i=1}^N \right)
\label{R}
\end{equation}
where $P_i$ denotes the spatial coordinates of the $i$-th masked block in the original image, and $\mathcal{S}(\cdot)$ is a spatial stitching operator that arranges each extracted patch $U_i$ according to $P_i$, preserving their relative spatial layout as in the original image.
This spatially-aware composition preserves relative positional consistency and complements the masked input by reintroducing critical visual information lost during masking.
% The image $R$ captures only the information removed in the main branch, offering a focused view for the model to learn from missing local content.
By reintroducing these semantically rich but commonly overlooked areas, the reuse branch not only complements the local representation, but also reduces the risk of information loss in standard masking-based learning.
% Although the mask image can serve as a regularization, it may result in the loss of key features, particularly in small target areas. Furthermore, the effectiveness of masking strongly depends on the mask strategy and mask ratio, which may not be universally suitable for all tasks or image distributions. To address these limitations, we design a mask information reuse branch, which is designed to re-learn and enhance the information lost due to masking.
% Specifically, first identify all the masked blocks within the input image and extract their corresponding unmasked regions from the original image. These patches are then reassembled in spatial position in order to construct a new image composed exclusively of the masked regions. This newly constructed image captures only the information removed in the main branch, offering a focused view for the model to learn from missing local content. By reintroducing these semantically rich but commonly overlooked areas, the reuse branch not only complements the local representation, but also reduces the risk of information loss in standard masking-based learning.

% To ensure a balanced representation of global and local information, we set the mask ratio at 25\%. This setting allows the primary masked image to retain sufficient global context, while the reuse branch focuses on refining missing local details. Further analysis of the impact of the mask ratio is presented in Section \ref{Mask Ratio}.

\subsection{Feature Fusion and Alignment}
Due to the size mismatch between the reuse image and the masked image, the reuse image is first upsampled to match the spatial dimensions of the masked image before being further processed. Then both images are input into a low-level feature extractor that shares the structure.
The extractor is constructed by selecting the shallow layers from a standard deep neural network backbone. These early layers retain high-resolution spatial information and are well-suited for capturing fine-grained local patterns, including edges, corners, and textures. The extracted features from both inputs are concatenated along the channel dimension to form a joint representation.

To address the semantic inconsistency between global and local features, we design a feature alignment module. This module performs soft alignment by applying a series (n, default n=3) of convolutional layers and preserves fine-grained details and local texture through a multi-level residual refinement mechanism. After alignment, the resulting feature is forwarded to a high-level feature extractor, which consists of the deeper layers of the same backbone network used in the low-level feature extractor. This design ensures hierarchical consistency while allowing the model to capture enriched semantic features across multiple scales.
It is precisely this modular decomposition that enables our method to be compatible with any backbone architecture and provides significant flexibility in design. Additionally, constraining the mask reuse branch to operate in the low-level feature extraction stage allows efficient control of additional parameters and computational cost.

% Furthermore, to ensure the diversity of training and the extent of domain shift, a hyperparameter $P$ is introduced to control the use of the mask probability. When the mask method is not applied, the main branch directly uses the original image for training, while the supplementary branch utilizes the original image with varying degrees of augmentation for training.

\section{Experiments and Results}
\subsection{Implementation Details}
\textbf{Basic Environment Settings}
All experiments were conducted on a computer with an Intel Core 17-10700 CPU, 64G of RAM and an NVIDIA RTX 4090 GPU. The equipped software runtime environment was also set up with Pycharm2024, python 3.9, PyTorch 2.1.1, CUDA 11.8 and cuDNN 8.6.0.

\textbf{Classification Model hyperparameters Settings}
All classification models used in the experiments, as well as the corresponding MaskAnyNet variants, are trained with the same hyperparameters as their original baseline implementations. Specifically, we train for 300 epochs on CIFAR datasets and 200 epochs on ImageNet.

\textbf{Downstream Model Hyperparameters Settings}
In order to ensure fair comparison of model performance in downstream task comparison experiments, we set hyperparameters for different baseline models, as shown in Table \ref{Hyperparameters Settings}. When training with MaskAnyNet, it inherits the same hyperparameter settings as the baseline. This ensures that any observed performance differences are attributed to the design of MaskAnyNet, rather than external training factors such as different learning rates or optimization strategies.

\begin{table}[!htbp]
\caption{Key hyperparameter configurations employed in different model training processes}
\label{Hyperparameters Settings}
\begin{tabular}{lcc}
\toprule
Model & Config & Value \\
\midrule
\multirow{12}{*}{\rotatebox{90}{RT-DETR}}
& Optimizer & SGD \\ 
& Base learning rate & 0.01 \\ 
& Final learning rate & 0.0001 \\
& Policy & cos \\
& Weight decay & 0.0005 \\
& Optimizer momentum & 0.937\\
& Warmup momentum & 0.8 \\
& Batch size & 32 \\ 
& Training epochs & 300 \\ 
& Warmup epochs & 3 \\
& Mosaic & 1.0 \\ 
& Dropout ratio & 0.0 \\
\hline
\multirow{12}{*}{\rotatebox{90}{YOLO12}}
& Optimizer & SGD \\ 
& Base learning rate & 0.01 \\ 
& Final learning rate & 0.0001 \\ 
& Policy & Cos \\
& Weight decay & 0.0005 \\
& Optimizer momentum & 0.937\\
& Warmup momentum & 0.8 \\
& Batch size & 32 \\ 
& Training epochs & 300 \\ 
& Warmup epochs & 3 \\
& Mosaic & 1.0 \\ 
& Dropout ratio & 0.0 \\
\hline
\multirow{12}{*}{\rotatebox{90}{Segformer}}
& Optimizer & AdamW \\ 
& Base learning rate & 0.00006 \\ 
& Final learning rate & 0.0 \\
& Policy & Poly \\
& Weight decay & 0.01 \\
& Batch size & 16 \\ 
& Iterations & 160K \\ 
& Warmup iters & 1500 \\
& Mosaic & 0.0 \\ 
& Dropout ratio & 0.1 \\
\hline
\multirow{12}{*}{\rotatebox{90}{DeepLabV3+}}
& Optimizer & Nesterov momentum \\ 
& Base learning rate & 0.007 \\ 
& Final learning rate & 0.0 \\
& Policy & Poly \\
& Weight decay & 0.00004 \\
& Optimizer momentum & 0.9\\
& Batch size & 32 \\ 
& Iterations & 160K \\ 
& Warmup iters & 1500 \\
& Mosaic & 0.0 \\ 
& Dropout ratio & 0.0 \\
\bottomrule
\end{tabular}
\end{table}

\subsection{Cifar-10 and Cifar-100 Experiments}
To evaluate the effectiveness and generalizability of our approach, we conducted comparative experiments on CIFAR-10 and CIFAR-100 using representative CNN and Transformer models. The comparison results are presented in Table \ref{Table4}. Specifically, we selected ResNet-34 and EfficientNet-V2 as CNN baseline models, with the former being a widely used architecture and the latter achieving the best performance among the CNN baselines. For Transformer-based baselines, we chose ViT and Swin, both of which are well-established and effective in vision tasks.
As shown in Table \ref{Table4}, our method improves the accuracy across all models on both CIFAR-10 and CIFAR-100. For example, ResNet-34 and EfficientNet-V2 gain 1.56\% and 1.03\% on CIFAR-10, respectively, while ViT and Swin improve by 0.42\% and 0.67\%. On CIFAR-100, the gains are even more pronounced, with up to 1.78\% on ResNet-34, 1.42\% on EfficientNet-V2, and 0.97\% on ViT, demonstrating the strong generalization ability of our method. The larger gains on the more challenging CIFAR-100 dataset further suggest that the proposed reuse strategy is particularly effective for fine-grained recognition tasks.

% To comprehensively evaluate the effectiveness and applicability of our approach, we performed comparative experiments across a variety of representative CNN-based and Transformer-based models. Among the CNNs, the ResNet-34 network, known for its representative performance, and EfficientNet-V2, which exhibited the best performance among all baseline models, were chosen as the primary benchmarks for comparison. For Transformer-based models, ViT and Swin, two widely adopted backbone architectures, were selected for evaluation. The comparison results are presented in Table \ref{Table4}.
% It can be observed that our method significantly improves the performance of the baseline models, particularly on the more challenging CIFAR-100 dataset. Ours method demonstrates strong generalization across both CNN and Transformer architectures, with notable improvements in Top-1 accuracy, achieving gains of 1.78\% on ResNet-34 and 0.97\% on Vit.

\setlength{\tabcolsep}{2pt} % 调整列间间距
\renewcommand{\arraystretch}{1} % 调整行间距
\begin{table}[ht]
\centering
\caption{Comparison of Top-1 Accuracy of MaskAnyNet and different models on Cifar-10 and Cifar-100 datasets.}
\label{Table4}
\begin{tabular}{llc}
\toprule
$\mathcal{D}$ & Method & Top-1 Acc \\ 
\midrule
\multirow{13}{*}{\rotatebox{90}{Cifar-10}} 
& ResNet-34\cite{A1} & 94.21 \\ 
& EfficientNetV2-S\cite{A2} & 94.72 \\ 
& MobileNetV4-Small\cite{A3} & 93.23 \\ 
& GhostNet-V3\cite{A4} & 94.12 \\ 
& FasterNet-S\cite{A5} & 91.95 \\ 
& ConvNextV2-Tiny\cite{r17} & 94.67 \\ 
% & Swin-Tiny\cite{A6} & 88.30 \\ 
& Swin-Tiny\cite{A6} & 96.56 \\ 
% & Vit-Base\cite{A7} & 65.55 \\ 
& Vit-Base\cite{A7} & 98.99 \\
& \cellcolor{gray!10}MaskResNet-34 & \cellcolor{gray!10}\textbf{(+1.56)95.77} \\
& \cellcolor{gray!10}MaskEfficientV2 & \cellcolor{gray!10}\textbf{(+1.03)95.75} \\
% & \cellcolor{gray!10}MaskViT & \cellcolor{gray!10}\textbf{(+0.89)66.44} \\
& \cellcolor{gray!10}MaskViT & \cellcolor{gray!10}\textbf{(+0.42)99.41} \\
% & \cellcolor{gray!10}MaskSwin & \cellcolor{gray!10}\textbf{(+0.42)88.72} \\
& \cellcolor{gray!10}MaskSwin & \cellcolor{gray!10}\textbf{(+0.67)97.23} \\ 
\hline
\multirow{13}{*}{\rotatebox{90}{Cifar-100}} 
& ResNet-34\cite{A1} & 73.58 \\ 
& EfficientNetV2-S\cite{A2} & 76.71 \\ 
& MobileNetV4-Small\cite{A3} & 71.78 \\ 
& GhostNet-V3\cite{A4} & 74.09 \\
& FasterNet-S\cite{A5} & 70.31 \\ 
& ConvNextV2-Tiny\cite{r17} & 72.32 \\ 
% & Swin-Tiny\cite{A6} & 57.74 \\ 
& Swin-Tiny\cite{A6} & 90.46 \\ 
% & Vit-Base\cite{A7} & 44.18 \\ 
& Vit-Base\cite{A7} & 91.90 \\
& \cellcolor{gray!10}MaskResNet-34 & \cellcolor{gray!10}\textbf{(+1.78)75.36} \\
& \cellcolor{gray!10}MaskEfficientV2 & \cellcolor{gray!10}\textbf{(+1.49)78.20} \\
% & \cellcolor{gray!10}MaskViT & \cellcolor{gray!10}\textbf{(+0.4)44.58} \\
& \cellcolor{gray!10}MaskViT & \cellcolor{gray!10}\textbf{(+0.97)92.87} \\
% & \cellcolor{gray!10}MaskSwin & \cellcolor{gray!10}\textbf{(+0.7)58.44} \\
& \cellcolor{gray!10}MaskSwin & \cellcolor{gray!10}\textbf{(+0.77)91.23} \\ 
\bottomrule
\end{tabular}
\end{table}

\subsection{ImageNet Experiments}
To further validate the universality and robustness of MaskAnyNet, we conducted experiments on the large-scale ImageNet-1K and Tiny-ImageNet datasets, as presented in Table \ref{ImageNet Experiments}. The results show consistent improvements across all baseline. Specifically, on ImageNet-1K, all baseline models achieve more than a 1.0\% improvement in Top-1 accuracy when integrated with our approach. Notably, the Top-1 accuracy of ViT-B/16 and ResNet-34 increases significantly by 1.56\% and 1.45\%, respectively.
Furthermore, the performance gains on Tiny-ImageNet are more pronounced. Specifically, our method improves the Top-1 accuracy of various models, including 2.08\% on ResNet-34, 1.28\% on EfficientNetV2, 1.68\% on ViT, and 1.71\% on Swin, demonstrating consistent gains across both CNN and Transformer architectures. Compared to ImageNet-1k, the limited amount of training data in Tiny-ImageNet causes the baselines to suffer from poor generalization. In contrast, our method not only provides a regularization effect, but also enhances pixel utilization by reusing masked information. This enables the model to learn more robust and diverse feature representations. As a result, our method significantly improves the performance of the baseline models.

Comparative results across multiple datasets confirm that our masked region reuse strategy generalizes well across different backbone architectures. By transforming discarded pixels into reusable learning signals, our method facilitates richer feature learning and improves the model’s ability to recognize small, occluded, or semantically diverse targets.
% To further validate the universality of our approach, additional experiments were conducted on the ImageNet dataset, with results presented in Table \ref{Table5}. The data reveal that even in more challenging scenarios, models incorporating our method, whether based on CNN or Transformer architectures, achieved significant improvements in both Top-1 and Top5 accuracy. This shows that our design paradigm effectively utilizes the same pixel resources to provide more diverse feature information, thereby enhancing model performance. Furthermore, the experimental results across CIFAR-10, CIFAR-100, and ImageNet datasets show that as the dataset size increases and tasks become more complex, the performance gains of models employing our method show no signs of saturation.

\renewcommand{\arraystretch}{1} % 调整行间距，默认值是1
\begin{table}[ht]
\centering
\caption{Comparison of Top-1 and Top-5 accuracy of MaskAnyNet and different models on ImageNet-1K and Tiny-ImageNet datasets.}
\label{ImageNet Experiments}
\resizebox{0.48\textwidth}{!}{
\begin{tabular}{llccc}
\toprule
$\mathcal{D}$ & Method & Top-1 Acc & Top-5 Acc \\ 
\midrule
\multirow{12}{*}{\rotatebox{90}{ImageNet-1K}}
 & ResNet-34\cite{A1}  & 71.67 & 90.64 \\ 
 & EfficientNetV2-S\cite{A2} & 74.73 & 91.25 \\ 
 & MobileNetV4-Small\cite{A3} & 68.95 & 89.62 \\ 
 & GhostNet-V3\cite{A4} & 69.58 & 89.31 \\ 
 & FasterNet-S\cite{A5} & 70.68 & 90.88 \\ 
 & ConvNextV2-Tiny\cite{r17} & 74.68 & 91.54 \\ 
 & Swin-T\cite{A6} & 78.63 & 93.87 \\ 
 & Vit-B/16\cite{A7} & 79.51 & 94.62 \\
 & \cellcolor{gray!10}MaskResNet-34 & \cellcolor{gray!10}\textbf{(+1.45)73.12} & \cellcolor{gray!10}\textbf{(+1.19)91.83} \\
 & \cellcolor{gray!10}MaskEfficientV2-S & \cellcolor{gray!10}\textbf{(+1.12)75.85} & \cellcolor{gray!10}\textbf{(+0.29)91.54} \\
 & \cellcolor{gray!10}MaskViT-B/16 & \cellcolor{gray!10}\textbf{(+1.56)81.07} & \cellcolor{gray!10}\textbf{(+0.76)95.38} \\
 & \cellcolor{gray!10}MaskSwin-T & \cellcolor{gray!10}\textbf{(+1.32)79.95} & \cellcolor{gray!10}\textbf{(+0.61)94.48} \\
\hline
\multirow{14}{*}{\rotatebox{90}{Tiny-ImageNet}} 
 & ResNet-34\cite{A1}  & 60.27 & 80.98 \\ 
 & EfficientNetV2-S\cite{A2} & 64.57 & 84.84 \\ 
 & MobileNetV4-Small\cite{A3} & 60.57 & 82.83 \\ 
 & GhostNet-V3\cite{A4} & 58.49 & 81.42 \\ 
 & FasterNet-S\cite{A5} & 58.67 & 84.32 \\ 
 & ConvNextV2-Tiny\cite{r17} & 59.05 & 79.04 \\ 
% Swin-T\cite{A6} & 46.94 & 70.06 \\ 
 & Swin-T\cite{A6} & 82.32 & 93.61 \\ 
% Vit-B\cite{A7} & 28.26 & 53.18 \\ 
 & Vit-B\cite{A7} & 85.83 & 95.45 \\
 & \cellcolor{gray!10}MaskResNet-34 & \cellcolor{gray!10}\textbf{(+2.08)62.35} & \cellcolor{gray!10}\textbf{(+1.28)82.26} \\
 & \cellcolor{gray!10}MaskEfficientV2 & \cellcolor{gray!10}\textbf{(+1.98)66.55} & \cellcolor{gray!10}\textbf{(+0.11)84.95} \\
% \rowcolor{gray!10} MaskViT & \textbf{(+0.88)29.81} & \textbf{(+2.29)55.47} \\
 & \cellcolor{gray!10}MaskViT & \cellcolor{gray!10}\textbf{(+1.68)87.51} & \cellcolor{gray!10}\textbf{(+0.96)96.41} \\
% \rowcolor{gray!10} MaskSwin & \textbf{(+1.36)48.30} & \textbf{(+2.26)72.32} \\
 & \cellcolor{gray!10}MaskSwin & \cellcolor{gray!10}\textbf{(+1.71)84.03} & \cellcolor{gray!10}\textbf{(+0.61)94.22} \\ 
 \bottomrule
\end{tabular}
}
\end{table}

% \setlength{\tabcolsep}{1pt} % 调整列间间距，默认值是6pt
% \renewcommand{\arraystretch}{1} % 调整行间距，默认值是1
% \begin{table}[htbp]
% \centering
% \caption{Comparison of Top-1 and Top5 Accuracy of MaskAnyNet and Different Models on Tiny-ImageNet Dataset}
% \label{Table5}
% \begin{tabular}{lccc}
% \toprule
% Method  & Dataset & Top-1\_Acc & Top5\_Acc \\ 
% \midrule
% ResNet-34\cite{A1}   &\multirow{16}{*}{\rotatebox{90}{Tiny-ImageNet}}     & 60.27 & 80.98 \\ 
% EfficientNetV2-S\cite{A2} &                   & 64.57 & 84.84 \\ 
% MobileNetV4-Small\cite{A3} &                  & 56.57 & 79.83 \\ 
% GhostNet-V3\cite{A4} &                        & 58.49 & 81.42 \\ 
% FasterNet-S\cite{A5} &                        & 58.67 & 84.32 \\ 
% ConvNextV2-Tiny\cite{r17} &                   & 59.05 & 79.04 \\ 
% Swin-T\cite{A6} &                             & 46.94 & 70.06 \\ 
% Swin-T(PreTrain) &                            & 78.32 & 93.61 \\ 
% Vit-B\cite{A7} &                              & 28.26 & 53.18 \\ 
% Vit-B(PreTrain) &                             & 85.83 & 95.45 \\ 
% MaskResNet-34 &                   & (+1.75)62.35 & (+1.28)82.68 \\ 
% MaskEfficientV2 &                 & (+1.25)65.82 & (+0.11)84.95 \\ 
% MaskViT &                         & (+0.88)29.81 & (+2.29)55.47 \\ 
% MaskViT(PreTrain) &               & (+1.68)87.85 & (+0.96)96.74 \\ 
% MaskSwin&                        & (+1.36)48.30 & (+2.26)72.32 \\ 
% MaskSwin(PreTrain) &              & (+1.71)80.03 & (+0.61)94.21 \\ 
% \bottomrule
% \end{tabular}
% \end{table}

\subsection{Ablation Experiments}
To investigate the contribution of each component in our framework, we conducted ablation studies on the ImageNet-1K dataset using both ResNet-34 and ViT-B/16 as baseline, as shown in Table \ref{Table6}.
Specifically, $\mathcal{M}$ denotes the use of mask augmentation, $\mathcal{R}$ indicates the mask region reuse branch, and $FFA$ refers to the feature fusion and alignment module.
\setlength{\tabcolsep}{4pt}
\renewcommand{\arraystretch}{1} % 调整行间距，默认值是1
\begin{table}[htbp]
\centering
\caption{Ablation experiment on the ImageNet-1k
dataset.}
\label{Table6}
\resizebox{0.46\textwidth}{!}{
\begin{tabular}{lccccc}
\toprule
Method  & $\mathcal{M}$ & $\mathcal{R}$ & $FFA$ & Top-1 Acc & Top-5 Acc\\ 
\midrule
RseNet-34 & - & - & - & 71.67  & 90.64\\ 
Patch Mask & \checkmark & - & - & 72.12 & 91.25 \\ 
Grid Mask & \checkmark & - & - & 71.97 & 91.14 \\
Random Mask & \checkmark & - & - & 71.88 & 91.02 \\ 
MaskResNet-34 & \checkmark & \checkmark & - & 72.97  & 91.76 \\
MaskResNet-34 & \checkmark & \checkmark & \checkmark & 73.12  & 91.83 \\ 
\hline
Vit-B/16   & - & - & - & 79.51 & 94.62 \\ 
Patch Mask & \checkmark & - & - & 79.88 & 94.78 \\ 
Grid Mask & \checkmark & - & - &  79.95 & 94.83 \\
Random Mask & \checkmark & - & - & 79.86 & 94.67 \\
MaskVit-Base & \checkmark & \checkmark & - & 81.01  & 95.23 \\ 
MaskVit-Base & \checkmark & \checkmark & \checkmark & 81.07 & 95.38 \\ 
\bottomrule
\end{tabular}
}
\end{table}
Introducing masking augmentation leads to certain improvements, demonstrating the regularization and local perturbation benefits of masking.
Further incorporating the reuse branch results in the most significant improvement, with Top-1 accuracy gains of 0.85\% for ResNet-34 and 1.06\% for ViT-B/16, confirming the effectiveness of reintroducing masked region information.
Finally, introducing $FFA$ results in the best performance, highlighting that it effectively bridges the semantic gap between both branches and improves the integration of global-local features.
\subsection{Masking Strategy Analysis}
\label{Masking Strategy}
We evaluated three masking strategies on the ImageNet-1K dataset using ResNet-34, and the results are shown in Table \ref{Table1}. As observed in the table, random masking yields the lowest performance, due to significant structural deviations in the reuse image, which hinder feature alignment between the masked and reused images and affect the model’s learning process.
Grid masking performs better by preserving global semantics. The reuse image generated through grid sampling effectively restores global information. However, its overall feature distribution closely resembles that of the masked image, which may cause the model to exhibit "lazy" behavior, relying too heavily on redundant information, thus limiting further performance improvement.
In contrast, patch masking achieves the highest accuracy by introducing local variability and preserving critical object details. It provides regularization benefits for the masked image and ensures that key features are retained through the reuse process. Moreover, this regional reuse strategy enables the model to capture fine-grained local patterns, enhancing representational capacity and generalization.
Notably, combining patch and grid masking achieves the best result, demonstrating the complementary effect of global structure and local detail reuse. Based on these findings, our final reuse strategy adopts both patch masks and grid masks.

% We further analyzed the usage probability of various mask reuse methods. Using different schemes that involve Random Mask, Grid Mask, and Patch Mask, we systematically evaluated the probabilities of their use to identify the optimal masking strategy. The results are summarized in the table.
\setlength{\tabcolsep}{7pt} % 调整列间间距，默认值是6pt
\renewcommand{\arraystretch}{1} % 调整行间距，默认值是1
\begin{table}[htbp]
\centering
\caption{Performance comparison on the ImageNet-1k using different mask and reuse methods based on ResNet-34.}
\label{Table1}
\begin{tabular}{lcccc}
\toprule
Patch  & Grid & Random & Top-1 Acc & Top-5 Acc\\ 
\midrule
\checkmark &             &            & 72.89     & 92.03 \\ 
           & \checkmark  &            & 72.72     & 91.98 \\ 
           &             & \checkmark & 72.23     & 91.73 \\ 
\cellcolor{gray!10}\checkmark & \cellcolor{gray!10}\checkmark  & \cellcolor{gray!10}           & \cellcolor{gray!10}\textbf{73.12}    & \cellcolor{gray!10}\textbf{91.83} \\ 
\checkmark & \checkmark  & \checkmark & 73.07     & 91.67 \\
\bottomrule
\end{tabular}
\end{table}

\subsection{Mask Ratio Analysis}
\label{Mask Ratio}
\begin{figure}[htbp]
\centering %表示居中
\includegraphics[width=8.5cm,height=6cm]{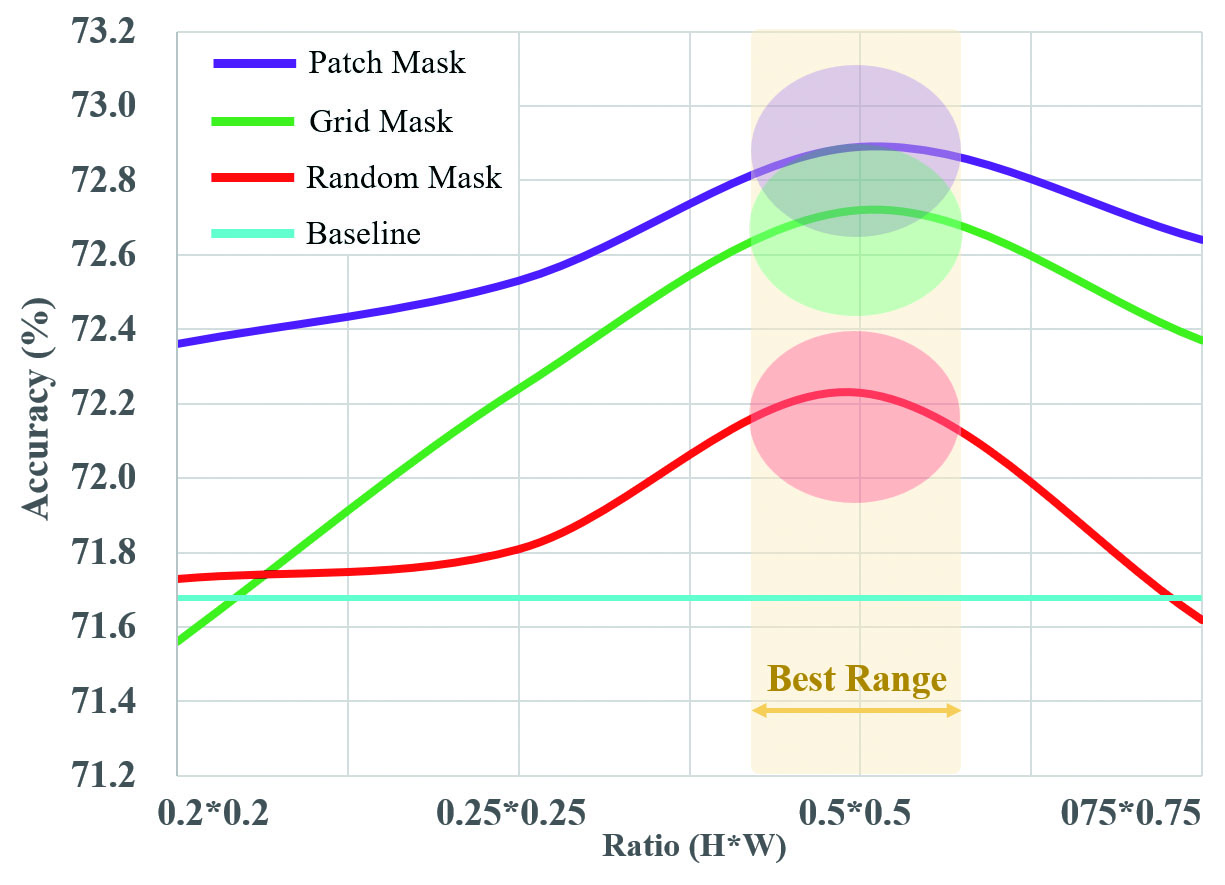}
\caption{Performance visualization results of MaskResNet-34 using three different masking methods and different mask ratios on the ImageNet-1K dataset.}
\label{Fig4}
\end{figure}

In addition to the choice of masking strategy, the mask ratio also plays a crucial role in determining how much visual information is retained, thereby directly affecting the model's ability to learn diverse representations. To investigate this, we evaluated different masking ratios across different masking methods, and the results are presented in Figure \ref{Fig4}.

% As shown in Figure \ref{Fig4}, a low mask ratio leads to weak regularization and insufficient information in the reused image, resulting in poor accuracy. As the mask ratio increases, the model accuracy improves rapidly. However, further increasing the mask ratio results in a decline in accuracy. Regardless of the mask method used, the peak accuracy is generally observed around a mask ratio of 25\%.

As shown in Figure \ref{Fig4}, a low mask ratio leads to weak regularization and insufficient information in the reused image, resulting in poor accuracy. As the ratio increases, the accuracy of all methods improves, reaching the best performance at a mask area ratio of 25\%. Further increasing the ratio leads to a decrease in accuracy, indicating that excessive occlusion limits the model’s to learn informative content.

\subsection{Feature Fusion Strategy Analysis}
To effectively integrate features from different branches, it is essential to choose an appropriate fusion location. The most common approaches include image-level fusion\cite{T8,T9}, feature-level fusion\cite{T10,T11}, and decision-level\cite{T12,T13} fusion. We evaluate these three strategies on the ImageNet-1K using ResNet and pre-trained ViT-B/16. Table \ref{Table2} shows the comparison of these three strategies in terms of accuracy and computational efficiency. It can be seen that the method of using feature fusion consistently achieves the highest accuracy in both ResNet and Vit, outperforming the other two methods. This proves that feature layer fusion can achieve good filtering of noise while aligning two branches information through remaining deep layers of the model, with high robustness.

In terms of computational efficiency, using low-level feature fusion can maintain a similar number of parameters as image-level fusion, and the computational burden introduced by the additional branch can be largely ignored. However, decision-level fusion nearly doubles the computational cost. Moreover, due to the depth of the fusion site, it becomes difficult to effectively align features from the two branches, resulting in limited improvement in accuracy.
% The following provides an analysis of the characteristics of these three fusion strategies.
% Image-level fusion directly combines input images, preserving full visual details with low computational cost. However, as it operates on raw pixels, it is prone to noise and may produce unstable results. Additionally, because it operates at the lowest fusion level, it may limit the model's generalization capability.
% Feature-level fusion first extracts low-level features (e.g., edges, textures) before fusion. It offers better noise resistance and more stable results. However, its effectiveness largely depends on the quality of extracted features and the specific stage at which fusion is applied, requiring extensive experimentation and design tuning.
% Decision-level fusion merges the final outputs from each branch. Although simple to implement, the lack of alignment between the feature information from different branches may lead to feature conflicts. Furthermore, the use of independent deep processing paths increases computational.
% We tested the three strategies on the Tiny-ImageNet dataset using ResNet and pretrain ViT backbones, with results shown in Table \ref{Table2}. It can be seen that the feature-level fusion strategies achieve the best performance on both the representative architectures of ViT and ResNet, demonstrating its strong robustness across different model designs.

\setlength{\tabcolsep}{5pt} % 调整列间间距，默认值是6pt
\renewcommand{\arraystretch}{1} % 调整行间距，默认值是1
\begin{table}[htbp]
\centering
\caption{Performance Comparison of Different Feature Fusion Strategies in Accuracy and Computational Efficiency.}
% Pa denotes the number of parameters, and T represents the average inference time per image.}
\label{Table2}
\resizebox{0.47\textwidth}{!}{
\begin{tabular}{lcccc}
\toprule
Model  & Fusion Level & Acc & Pa($M$) & T($per/ms$)\\ 
\midrule
\multirow{3}{*}{ResNet-34} &Image-level & 73.03 & 21.34 & 7.29 \\
 & Feature-level & 73.12 & 22.46 & 7.51 \\ 
 & Decision-level & 72.28 & 43.10 & 14.04 \\
\hline
\multirow{3}{*}{Vit-Base} & Image-level & 79.37 & 85.72 & 7.94 \\
 & Feature-level & 79.51 & 87.46 & 8.23 \\
 & Decision-level & 78.82 & 171.45 & 16.21 \\
\bottomrule
\end{tabular}
}
\end{table}

% \setlength{\tabcolsep}{5pt} % 调整列间间距，默认值是6pt
% \renewcommand{\arraystretch}{1} % 调整行间距，默认值是1
% \begin{table}[htbp]
% \centering
% \caption{The results of different fusion methods}
% \label{Table3}
% \begin{tabular}{lccc}
% \toprule
% Basic Model  & Fusion Strategy & Top-1 Acc & TOP5 Acc\\ 
% \midrule
% \multirow{2}{*}{ResNet-34}   & Add & 74.69 & 92.30 \\ 
%  & \cellcolor{gray!10}Concat & \cellcolor{gray!10}\textbf{75.36} & \cellcolor{gray!10}\textbf{92.44}  \\
% \hline
% \multirow{2}{*}{Vit-Base}   & Add & 43.90 & 73.30 \\ 
% & \cellcolor{gray!10}Concat & \cellcolor{gray!10}\textbf{44.58} & \cellcolor{gray!10}\textbf{73.80}  \\
% \bottomrule
% \end{tabular}
% \end{table}

% \subsection{Training schedule}
% The hyperparameters used during model training and the corresponding configurations are described in the supplementary material

\section{Detection and Segmentation Experiments}
Our method demonstrates notable performance gains on classification tasks, reflecting its strong feature representation capabilities. To further evaluate the generalizability of the proposed approach across downstream tasks, we conduct additional experiments on object detection and semantic segmentation, using representative CNN-based and Transformer-based models for comparison.

\subsection{Object detection}
In the object detection task, we employed YOLO12\cite{T2} (a CNN-based detection model) and RT-DETR\cite{T1} (a Transformer-based detection model) as benchmark models, testing on two authoritative datasets PASCAL VOC\cite{T3} and MS COCO\cite{T4}. Following \cite{Appendix1}, the training and testing of the PASCAL VOC dataset were carried out by using 5k images from VOC 2007 and 16k images from VOC 2012 (“07+12”) for training, while the VOC 2007 test set was utilized for evaluation. The training hyperparameters of YOLO12 and RT-DETR are consistent with those presented in Table \ref{Hyperparameters Settings}.

We used P, R, mAP@.5, and mAP@[.5,.95] as the primary evaluation metrics, and the final comparison results are shown in Table \ref{Object detection}. According to these experiments, our method achieved accuracy gains in both YOLO12 and RT-DETR. Notably, YOLO12-n exhibited 1.68\% and 1.89\% improvements in mAP@.5 on VOC and COCO, respectively; meanwhile, RT-DETR showed 1.32\% and 1.42\% improvements under the same metrics. It is worth emphasizing that our method maintained robust detection performance for objects spanning a wide range of scales, indicating that the mask information reuse mechanism effectively enhances the cross-scale representational capacity of the model, thus alleviating the missed or false detections of targets with varying sizes.

\begin{table*}[h]
\centering
\caption{Object detection improvements on MSCOCO and PASCAL VOC using RT-Dter and YOLO12}
\label{Object detection}
\resizebox{0.85\textwidth}{!}{
\begin{tabular}{llcccc}
\toprule
Dataset & Model & $Precision$ & $Recall$ & $mAP@.5$ & $mAP@[.5,.95]$ \\ 
\midrule
\multirow{4}{*}{\rotatebox{90}{VOC}} & RT-DETR-n  & 82.46 & 75.52 & 81.63 & 62.53 \\ 
& \cellcolor{gray!10}MaskRT-DETR-n & \cellcolor{gray!10}\textbf{(+0.57)83.03} & \cellcolor{gray!10}\textbf{(+0.76)76.28} & \cellcolor{gray!10}\textbf{(+1.32)82.95} & \cellcolor{gray!10}\textbf{(+0.85)63.38}\\
& YOLO12-n & 80.29 & 74.18 & 81.19 & 61.69\\ 
& \cellcolor{gray!10}MaskYOLO12-n & \cellcolor{gray!10}\textbf{(+1.02)81.31} & \cellcolor{gray!10}\textbf{(+0.75)74.93} & \cellcolor{gray!10}\textbf{(+1.68)82.87} & \cellcolor{gray!10}\textbf{(+0.93)62.62}\\ 
\hline
\multirow{4}{*}{\rotatebox{90}{COCO}} & RT-DETR-n  & 68.27 & 53.93 & 63.83 & 44.82 \\ 
& \cellcolor{gray!10}MaskRT-DETR-n & \cellcolor{gray!10}\textbf{(+0.92)69.19} & \cellcolor{gray!10}\textbf{(+0.74)54.67} & \cellcolor{gray!10}\textbf{(+1.42)65.25} & \cellcolor{gray!10}\textbf{(+1.04)45.86}\\
& YOLO12-n   & 65.51 & 49.58 & 54.46 & 38.86\\ 
& \cellcolor{gray!10}MaskYOLO12-n & \cellcolor{gray!10}\textbf{(+1.33)66.84} & \cellcolor{gray!10}\textbf{(+1.07)50.65} & \cellcolor{gray!10}\textbf{(+1.89)56.35} & \cellcolor{gray!10}\textbf{(+1.03)39.89}\\ 
\bottomrule
\end{tabular}
}
\end{table*}

\subsection{Semantic Segmentation}
Following the training setup of the ADE20K semantic segmentation task, we train SegFormer \cite{T6} and DeepLabV3+ \cite{T7} as baseline models.

For semantic segmentation task, we train DeepLabV3+\cite{T7} (CNN-based) and Segformer\cite{T6} (Transformer-based) on ADE20K  dataset\cite{T5} to evaluate the performance improvements offered by our method in high-granularity segmentation tasks. Both lightweight and deeper backbones were tested, and using mIoU as the primary evaluation metric, the results are shown in the Table \ref{Semantic Segmentation}.
It can be observed that, regardless of the baseline method, all the models incorporating our approach achieve better performance. DeepLabV3+ improves its mIoU by 1.83\% when using a lightweight backbone, while still achieving a gain of 1. 22\% with the deeper ResNet-101 backbone. This phenomenon suggests that lightweight networks possess greater room for improvemen, an observation also supported by Segformer results, which achieves a 0.71\% increase in mIoU on the deeper B4 backbone and a 1.28\% increase on the lighter B0 backbone.
Through both detection and segmentation experiments, it becomes clear that the proposed method demonstrates strong generalizability across diverse encoder architectures. In particular, it shows notable advantages on lightweight backbones, underscoring its effectiveness in balancing computational efficiency and accuracy.

\begin{table}[h]
\centering
\caption{Semantic segmentation improvements on ADE-20K using DeepLabV3+ and Segformer as baseline.}
\label{Semantic Segmentation}
\resizebox{0.48\textwidth}{!}{
\begin{tabular}{lccc}
\toprule
Model & $Encoder$ & $mIoU$ \\ 
\midrule
\multirow{2}{*}{{DeepLabV3+}} & MobileNetV2 & 34.00 \\ 
& ResNet-101 & 44.10 \\

\multirow{2}{*}{{MaskDeepLabV3+}} & \cellcolor{gray!10}MobileNetV2 & \cellcolor{gray!10}\textbf{(+1.83)35.83} \\ 
 & \cellcolor{gray!10}ResNet-101 & \cellcolor{gray!10}\textbf{(+1.22)45.32} \\
 
\multirow{2}{*}{{Segformer}}   & MiT-B0 & 37.40 \\ 
 & MiT-B4 &  51.10 \\

\multirow{2}{*}{{MaskSegformer}} & \cellcolor{gray!10}MiT-B0 & \cellcolor{gray!10}\textbf{(+1.28)38.68} \\
 & \cellcolor{gray!10}MiT-B4 & \cellcolor{gray!10}\textbf{(+0.71)51.81} \\
\bottomrule
\end{tabular}
}
\end{table}

\section{Further Analysis}
\subsection{Weakly-Localization}
\begin{figure*}[h]
    \centering
    \begin{subfigure}[b]{1.0\textwidth}
        \includegraphics[width=\linewidth]{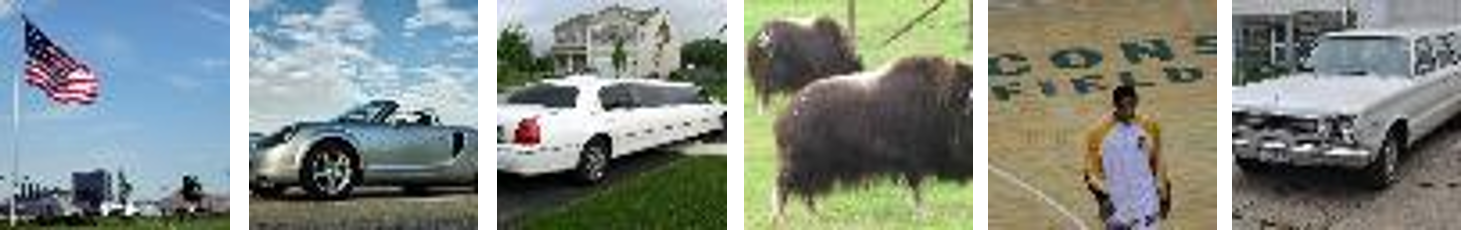}
        \caption{Origin Image}
        \label{Origin Image}
    \end{subfigure}%
    \par
        \begin{subfigure}[b]{1.0\textwidth}
        \includegraphics[width=\linewidth]{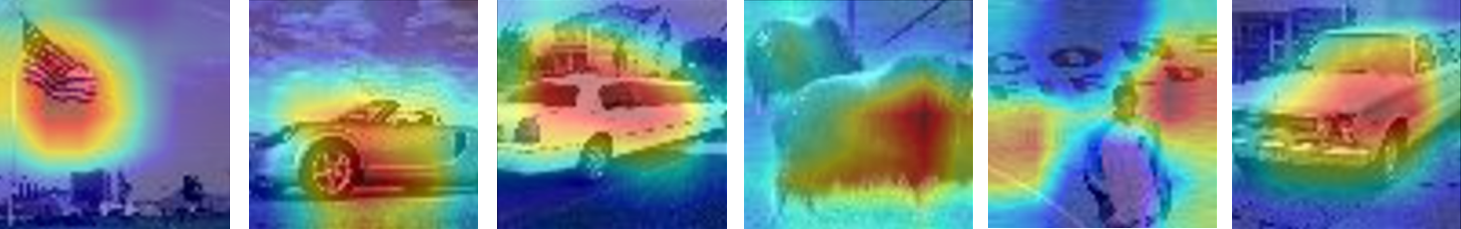}
        \caption{ResNet-34  Heat map}
        \label{ResNet-34 Heat map}
    \end{subfigure}%
    \par
        \begin{subfigure}[b]{1.0\textwidth}
        \includegraphics[width=\linewidth]{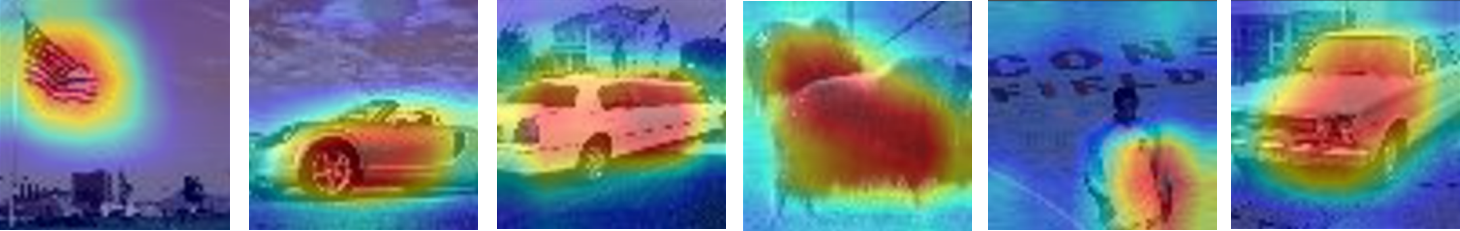}
        \caption{MaskResNet-34  Heat map}
        \label{MaskResNet-34  Heat map}
    \end{subfigure}%
    \caption{Grad-CAM Visualization of ResNet-34 and MaskResNet-34. It is evident that MaskResNet-34 focuses more accurately on the target regions (e.g., vehicle bodies, animal contours, human figures), while better suppressing background interference.}
    \label{Heat map}
\end{figure*}

We conducted weakly supervised heatmap analysis experiments on the Tiny ImageNet test dataset, visually comparing the decision-making basis of the ResNet-34 and MaskResNet-34 models using Grad-CAM\cite{Appendix2}. The results are presented in Figure \ref{Heat map}. As shown in the figure, traditional models, such as ResNet-34, are prone to background information interference (e.g. misidentification of humans and excessive focus on sky regions in flag recognition). They also tend to over-rely on local structures while lacking a global perception of the overall contour of the target (for example, recognizing only one animal while ignoring others in the image). In contrast, the model designed with MaskAnyNet demonstrates more precise attention mechanisms, effectively recognizing multiple targets (e.g., distinguishing multiple animals in the image) while preserving the overall contour of the target (e.g., capturing the complete shape of a vehicle). Additionally, background noise is significantly suppressed (e.g., reducing distractions from background text in human recognition and sky regions in flag recognition), and the heatmap activation areas align closely with the target semantics. These results validate the effectiveness of MaskAnyNet from an interpretability perspective. While retaining the regularization benefits of traditional mask-based methods, the proposed approach systematically integrates local details and global semantics through the collaborative training of mask region information recombination branches, effectively mitigates the issue of small target feature loss caused by random occlusion in conventional masking techniques. Heatmap analysis provides an intuitive attribution basis for the observed performance improvements, further demonstrating the effectiveness of the model in enhancing the robustness of feature representation.

\subsection{Image Similarity and Information Entropy}
\label{Image Similarity and Information Entropy}
Although the performance improvement achieved by our method has been quantitatively demonstrated, it remains unclear why different masking strategies lead to varying performance gains, particularly in terms of semantic differences in the information retained or reintroduced by each strategy. 
% Conventional metrics, such as classification accuracy, are insufficient to reveal the underlying dynamics of the reused information.
To investigate this, we analyze it from two perspectives: information diversity and information reliability. Information diversity is quantified using shannon entropy, which reflects the amount of uncertainty or richness in the image content\cite{F1}. In addition, the reliability of the information is assessed by deep feature similarity, indicating how semantically aligned the reused image is with the original image\cite{F2}. This analysis reveals how the reuse branch contributes to effective feature learning under different masking strategies.
Eq.\ref{Shannon entropy} defines the shannon entropy of the image $I$.
\begin{equation}
\label{Shannon entropy}
H(I) = -\sum_{i=0}^{255} p(i) \log_2 p(i)
\end{equation}
where $p(i)$ denotes the normalized frequency of the pixel value $i$ in the image. A higher entropy value indicates greater information diversity. The difference in entropy between the masked image and the reused image ($\Delta H$) reflects the amount of new information introduced by the reuse branch. Table \ref{Table7} shows the differences in the information entropy between different masked images and reused images.

\setlength{\tabcolsep}{16pt} % 调整列间间距，默认值是6pt
\renewcommand{\arraystretch}{1} % 调整行间距，默认值是1
\begin{table}[htbp]
\centering
\caption{Comparison of Information Entropy of Different Masking Methods}
\label{Table7}
\begin{tabular}{lccc}
\toprule
Method & $H_m$ & $H_c$ & $\Delta H$ \\ 
\midrule
Patch  & 6.224 & 7.147 & 0.923 \\ 
Grid   & 6.307 & 7.243 & 0.936 \\ 
Random & 6.297 & 6.911 & 0.614 \\ 
\bottomrule
\end{tabular}
\end{table}

As shown in Table \ref{Table7}, the Patch and Grid methods lead to larger entropy gains, suggesting that they introduce more diverse content into the learning process. In contrast, the Random method results in a smaller entropy difference, indicating limited additional information, which may explain its relatively lower effectiveness.

However, entropy alone does not capture the semantic quality of the reused information. For example, a large difference in information entropy may still be observed even if the reused image undergoes a significant domain shift, but the information may lack reliability.
% Additionally, when the reuse image closely resembles the original image, the richness of the introduced information may be limited.
To complement this, we evaluate the deep feature similarity. Specifically, we use a pre-trained ResNet-101 model to extract high-level features from both the original and reused images, and compute the cosine similarity. This result is transformed into a similarity score using Eq.\ref{formula2}, with the corresponding results summarized in Table \ref{Table8}.

\begin{equation}
\label{formula2}
S\left(I, I_c\right)=e^{-\left|S_{\text{ds}}\left(I, I_c\right)-S_{\text{a}}\right|}
\end{equation}
where $S_{\text{ds}}$ represents the similarity score between the reused image and the original image based on deep features, while $S_{\text{a}}$ is a hyperparameter that defines the optimal similarity value for deep features, which is set to 0.5 by default. 
To quantify this trade-off, we define a evaluation metric that combines entropy difference and similarity, defined as Eq.\ref{F score}
\begin{equation}
\label{F score}
F\left(I, I_m, I_c\right)=\frac{1}{N} \sum_{i=1}^N w_1 S\left(I, I_c\right)+w_2 \Delta H\left(I_m, I_c\right)
\end{equation}
where $N$ represents the number of sample pairs, which is set to 1000 in this case. $I$, $I_1^m$ and $I_c^i$ represent the original image, the masked image, and the reuse image, respectively. $w_1$ and $w_2$ represent different weight values, both set to 0.5 by default. The final results are presented in Table \ref{Table8}.

\setlength{\tabcolsep}{12pt} % 调整列间间距，默认值是6pt
\renewcommand{\arraystretch}{1} % 调整行间距，默认值是1
\begin{table}[htbp]
\centering
\caption{Similarity scores and final $\mathcal{F}$ scores obtained using Different Masking methods}
\label{Table8}
\begin{tabular}{lccc}
\toprule
Method & Model       & $S_{\text{ds}}$ & $\mathcal{F}$\\ 
\midrule
Patch  & \multirow{3}{*}{ResNet-101} & 0.537 & 0.943\\ 
Grid   &                             & 0.706 & 0.920\\ 
Random &                             & 0.426 & 0.771\\ 
\bottomrule
\end{tabular}
\end{table}
%As shown in Table \ref{Table8}, the Grid method exhibits the highest similarity due to its fixed sampling pattern, which may introduce redundant structural information and reduce feature sensitivity. In contrast, the Random method shows low similarity, which may cause feature conflicts and hinder effective learning. The Patch method achieves moderate similarity, striking a balance between diversity and reliability, leading to the highest final score $\mathcal{F}$.

As shown in Table \ref{Table8}, the Grid method achieves the highest similarity due to its fixed sampling pattern, but introduces structural redundancy that reduces feature sensitivity. Conversely, the Random method yields low similarity, causing feature conflicts that impede learning. The Patch method attains moderate similarity through optimal diversity-reliability trade-offs, achieving superior performance (highest $\mathcal{F}$).

\begin{figure}[htbp]
\centering %表示居中
\includegraphics[width=8.4cm,height=2.7cm]{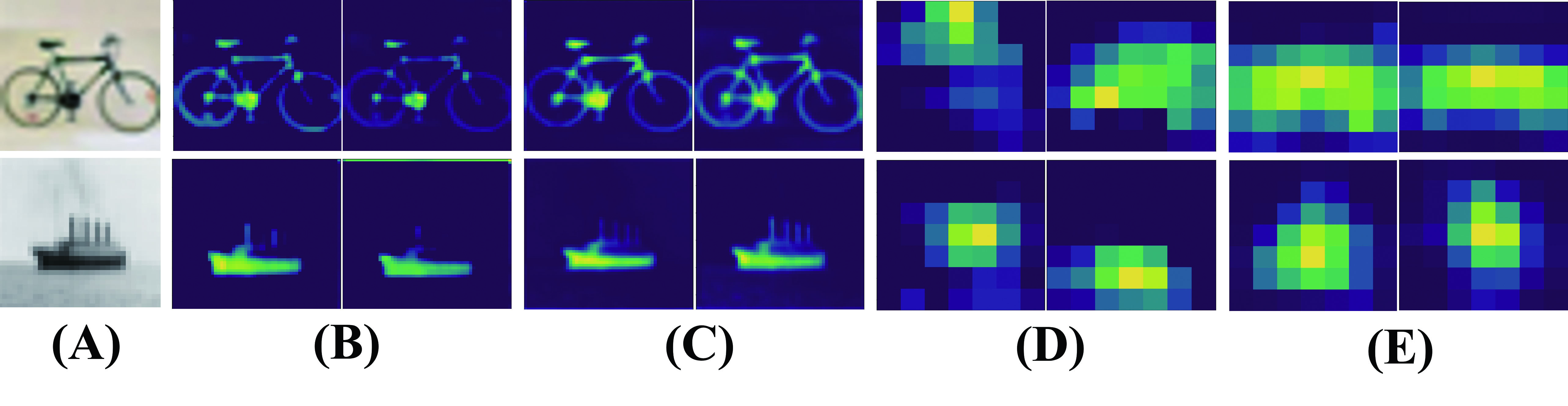}
\caption{Feature visualization comparisons of ResNet-34 and our MaskResNet-34. (A) Original image, (B) Low-level features from ResNet-34, (C) Low-level features from MaskResNet-34, (D) High-level features from ResNet-34, (E) High-level features from MaskResNet-34. 
% MaskResNet-34 preserves more object detail information in the low level features and locates more accurately in the high level features.
}
\label{Fig5}
\end{figure}

\subsection{Feature Visualization}
To intuitively understand the effectiveness of the proposed method, we visualize the intermediate feature representations of the baseline model (ResNet-34) and our model. As shown in Figure \ref{Fig5}, ResNet-34 tends to lose fine-grained structural details in the shallow layers (e.g. the chimney of the ship), causing blurred boundaries and imprecise localization in deeper layers.
In contrast, MaskResNet-34 incorporates a dual-branch design with an information reuse mechanism, enhancing the model’s sensitivity to local structures. This allows for better preservation of object boundaries and textures in early stages. In deeper layers, MaskAnyNet-34 produces more semantically aligned and spatially precise activations, indicating improved understanding of both object location and shape.
% These results confirm that the proposed strategy effectively enriches feature representations across the network hierarchy.

These observations confirm that our method not only strengthens local detail retention, but also promotes more robust global-semantic understanding, thereby enriching feature representations throughout the entire network hierarchy.

\section{Recommendations for the Application of the proposed MaskAnyNet}
Based on the extensive analysis and experiments for different tasks, we recommend the following points when applying the proposed MaskAnyNet.
\begin{itemize}
\item \textbf{Standard model and dataset.} If the same model and dataset as in this work are used, adopting the same settings is recommended to minimize parameter-tuning overhead.
\item \textbf{Novel model or Novel dataset.} For new models or datasets, first assess the size distribution of objects. If large objects dominate, the proportion of grid masking should be increased to broaden the receptive field and enhance inter-pixel interaction. Conversely, if smaller objects prevail, raise the proportion of patch masking to better preserve small-target details. For highly complex datasets or models, adjust the activation probability accordingly. These guidelines serve as references rather than absolute rules, and actual experimental results should guide final parameter tuning.
\end{itemize}

\section{Limitations and Future Learning}
\textbf{Limitations;} Despite the strong performance and generalizability demonstrated by MaskAnyNet across classification, detection, and segmentation tasks, several limitations remain that provide directions for future research. First, the introduction of a masked region reuse mechanism inevitably increases the number of hyperparameters, such as masking strategy, mask ratio, and feature fusion position. This makes the model more sensitive to configuration and may require extensive hyperparameter tuning to achieve optimal performance. Additionally, although the auxiliary branch components are lightweight, they introduce extra design and integration complexity when deploying.

\textbf{Future Learning;} In future work, we plan to (i) investigate adaptive masking strategies that dynamically adjust mask patterns based on dataset characteristics, enabling more context-aware information selection during training. (ii) to reduce the computational overhead of the reuse branch and make the model more suitable for real-time scenarios. and (iii) explore multimodal, semantic-aware reuse frameworks to further enhance model robustness and interpretability in complex real-world scenarios.

\section{Conclusions}
In this paper, we propose MaskAnyNet, a general visual architecture that addresses the limitations of traditional masking strategies in supervised learning, which commonly discard masked regions as noise, forfeiting fine-grained details. The proposed MaskAnyNet introduces a reuse branch to re-learn masked content and fuse it with the masked features, enhancing structural detail and feature diversity. This dual-branch design captures complementary global semantics and local intricacies, thereby significantly improving the generalization capabilities of the model. Extensive experiments on multiple benchmarks demonstrate consistent performance gains over state-of-the-art baselines. Further analysis highlights the importance of masking and fusion strategies, providing insight for future visual representation learning.

\section{Acknowledgments}
This work was supported in part by the National Natural Science Foundation of China (Grant Nos. 62373324, U24A20242, 62475241 and 62271448), in part by the Zhejiang Province Leading Geese Plan (Grant No. 2025C02160), in part by the National Key Research and Development Program of China (Grant No. 2024YFC3306902), in part by the Zhejiang Provincial Natural Science Foundation [grant no. LDT23F02024F02], and in part by Commercial Research Funds (Grant Nos. KYY-HX-20240822, KYY-HX-20200650, and KYY-ZH-20240057).

% \bigskip
% \noindent Thank you for reading these instructions carefully. We look forward to receiving your electronic files!

\bibliography{aaai2026}

@inproceedings{r1,
  title={Object region mining with adversarial erasing: A simple classification to semantic segmentation approach},
  author={Wei, Yunchao and Feng, Jiashi and Liang, Xiaodan and Cheng, Ming-Ming and Zhao, Yao and Yan, Shuicheng},
  booktitle={Proceedings of the IEEE conference on computer vision and pattern recognition},
  pages={1568--1576},
  year={2017}
}

@inproceedings{r2,
  title={Blockout: Dynamic model selection for hierarchical deep networks},
  author={Murdock, Calvin and Li, Zhen and Zhou, Howard and Duerig, Tom},
  booktitle={Proceedings of the IEEE conference on computer vision and pattern recognition},
  pages={2583--2591},
  year={2016}
}

@article{r3,
  title={Improved Regularization of Convolutional Neural Networks with Cutout},
  author={DeVries, Terrance},
  journal={arXiv preprint arXiv:1708.04552},
  year={2017}
}

@inproceedings{r4,
  title={Random erasing data augmentation},
  author={Zhong, Zhun and Zheng, Liang and Kang, Guoliang and Li, Shaozi and Yang, Yi},
  booktitle={Proceedings of the AAAI conference on artificial intelligence},
  volume={34},
  number={07},
  pages={13001--13008},
  year={2020}
}

@inproceedings{r5,
  title={Hide-and-seek: Forcing a network to be meticulous for weakly-supervised object and action localization},
  author={Kumar Singh, Krishna and Jae Lee, Yong},
  booktitle={Proceedings of the IEEE international conference on computer vision},
  pages={3524--3533},
  year={2017}
}

@article{r6,
  title={Gridmask data augmentation},
  author={Chen, Pengguang and Liu, Shu and Zhao, Hengshuang and Wang, Xingquan and Jia, Jiaya},
  journal={arXiv preprint arXiv:2001.04086},
  year={2020}
}

@article{r7,
  title={Fencemask: a data augmentation approach for pre-extracted image features},
  author={Li, Pu and Li, Xiangyang and Long, Xiang},
  journal={arXiv preprint arXiv:2006.07877},
  year={2020}
}

@INPROCEEDINGS{r8,
  author={Zhan, Yucheng and Zhao, Yucheng and Luo, Chong and Zhang, Yueyi and Sun, Xiaoyan},
  booktitle={2023 IEEE International Conference on Image Processing (ICIP)}, 
  title={Attention-Guided Contrastive Masked Image Modeling for Transformer-Based Self-Supervised Learning}, 
  year={2023},
  volume={},
  number={},
  pages={2490-2494},
  doi={10.1109/ICIP49359.2023.10222606}}

@article{r9,
  title={Beit: Bert pre-training of image transformers},
  author={Bao, Hangbo and Dong, Li and Piao, Songhao and Wei, Furu},
  journal={arXiv preprint arXiv:2106.08254},
  year={2021}
}

@article{r10,
  title={Bert: Pre-training of deep bidirectional transformers for language understanding},
  author={Devlin, Jacob},
  journal={arXiv preprint arXiv:1810.04805},
  year={2018}
}

@article{r11,
  title={Beit v2: Masked image modeling with vector-quantized visual tokenizers},
  author={Peng, Zhiliang and Dong, Li and Bao, Hangbo and Ye, Qixiang and Wei, Furu},
  journal={arXiv preprint arXiv:2208.06366},
  year={2022}
}

@inproceedings{r12,
  title={Masked autoencoders are scalable vision learners},
  author={He, Kaiming and Chen, Xinlei and Xie, Saining and Li, Yanghao and Doll{\'a}r, Piotr and Girshick, Ross},
  booktitle={Proceedings of the IEEE/CVF conference on computer vision and pattern recognition},
  pages={16000--16009},
  year={2022}
}

@inproceedings{r13,
  title={Simmim: A simple framework for masked image modeling},
  author={Xie, Zhenda and Zhang, Zheng and Cao, Yue and Lin, Yutong and Bao, Jianmin and Yao, Zhuliang and Dai, Qi and Hu, Han},
  booktitle={Proceedings of the IEEE/CVF conference on computer vision and pattern recognition},
  pages={9653--9663},
  year={2022}
}

@inproceedings{r14,
  title={Omnimae: Single model masked pretraining on images and videos},
  author={Girdhar, Rohit and El-Nouby, Alaaeldin and Singh, Mannat and Alwala, Kalyan Vasudev and Joulin, Armand and Misra, Ishan},
  booktitle={Proceedings of the IEEE/CVF conference on computer vision and pattern recognition},
  pages={10406--10417},
  year={2023}
}

@article{r15,
  title={Pixmim: Rethinking pixel reconstruction in masked image modeling},
  author={Liu, Yuan and Zhang, Songyang and Chen, Jiacheng and Chen, Kai and Lin, Dahua},
  journal={arXiv preprint arXiv:2303.02416},
  year={2023}
}

@article{r16,
  title={MixMask: Revisiting Masking Strategy for Siamese ConvNets},
  author={Vishniakov, Kirill and Xing, Eric and Shen, Zhiqiang},
  journal={arXiv preprint arXiv:2210.11456},
  year={2022}
}

@inproceedings{r17,
  title={Convnext v2: Co-designing and scaling convnets with masked autoencoders},
  author={Woo, Sanghyun and Debnath, Shoubhik and Hu, Ronghang and Chen, Xinlei and Liu, Zhuang and Kweon, In So and Xie, Saining},
  booktitle={Proceedings of the IEEE/CVF Conference on Computer Vision and Pattern Recognition},
  pages={16133--16142},
  year={2023}
}

@inproceedings{R18,
  title={Restore anything with masks: Leveraging mask image modeling for blind all-in-one image restoration},
  author={Qin, Chu-Jie and Wu, Rui-Qi and Liu, Zikun and Lin, Xin and Guo, Chun-Le and Park, Hyun Hee and Li, Chongyi},
  booktitle={European Conference on Computer Vision},
  pages={364--380},
  year={2024},
  organization={Springer}
}

@article{R19,
  title={Context autoencoder for self-supervised representation learning},
  author={Chen, Xiaokang and Ding, Mingyu and Wang, Xiaodi and Xin, Ying and Mo, Shentong and Wang, Yunhao and Han, Shumin and Luo, Ping and Zeng, Gang and Wang, Jingdong},
  journal={International Journal of Computer Vision},
  volume={132},
  number={1},
  pages={208--223},
  year={2024},
  publisher={Springer}
}

@inproceedings{R20,
  title={Masked feature prediction for self-supervised visual pre-training},
  author={Wei, Chen and Fan, Haoqi and Xie, Saining and Wu, Chao-Yuan and Yuille, Alan and Feichtenhofer, Christoph},
  booktitle={Proceedings of the IEEE/CVF conference on computer vision and pattern recognition},
  pages={14668--14678},
  year={2022}
}

@inproceedings{R21,
  title={Peco: Perceptual codebook for bert pre-training of vision transformers},
  author={Dong, Xiaoyi and Bao, Jianmin and Zhang, Ting and Chen, Dongdong and Zhang, Weiming and Yuan, Lu and Chen, Dong and Wen, Fang and Yu, Nenghai and Guo, Baining},
  booktitle={Proceedings of the AAAI conference on artificial intelligence},
  volume={37},
  number={1},
  pages={552--560},
  year={2023}
}

@inproceedings{R22,
  title={A-fast-rcnn: Hard positive generation via adversary for object detection},
  author={Wang, Xiaolong and Shrivastava, Abhinav and Gupta, Abhinav},
  booktitle={Proceedings of the IEEE conference on computer vision and pattern recognition},
  pages={2606--2615},
  year={2017}
}

@article{R23,
  title={Dropblock: A regularization method for convolutional networks},
  author={Ghiasi, Golnaz and Lin, Tsung-Yi and Le, Quoc V},
  journal={Advances in neural information processing systems},
  volume={31},
  year={2018}
}

@inproceedings{A1,
  title={Deep residual learning for image recognition},
  author={He, Kaiming and Zhang, Xiangyu and Ren, Shaoqing and Sun, Jian},
  booktitle={Proceedings of the IEEE conference on computer vision and pattern recognition},
  pages={770--778},
  year={2016}
}

@inproceedings{A2,
  title={Efficientnetv2: Smaller models and faster training},
  author={Tan, Mingxing and Le, Quoc},
  booktitle={International conference on machine learning},
  pages={10096--10106},
  year={2021},
  organization={PMLR}
}

@inproceedings{A3,
  title={MobileNetV4: Universal Models for the Mobile Ecosystem},
  author={Qin, Danfeng and Leichner, Chas and Delakis, Manolis and Fornoni, Marco and Luo, Shixin and Yang, Fan and Wang, Weijun and Banbury, Colby and Ye, Chengxi and Akin, Berkin and others},
  booktitle={European Conference on Computer Vision},
  pages={78--96},
  year={2025},
  organization={Springer}
}

@article{A4,
  title={GhostNetV3: Exploring the Training Strategies for Compact Models},
  author={Liu, Zhenhua and Hao, Zhiwei and Han, Kai and Tang, Yehui and Wang, Yunhe},
  journal={arXiv preprint arXiv:2404.11202},
  year={2024}
}

@inproceedings{A5,
  title={Run, don't walk: chasing higher FLOPS for faster neural networks},
  author={Chen, Jierun and Kao, Shiu-hong and He, Hao and Zhuo, Weipeng and Wen, Song and Lee, Chul-Ho and Chan, S-H Gary},
  booktitle={Proceedings of the IEEE/CVF conference on computer vision and pattern recognition},
  pages={12021--12031},
  year={2023}
}

@inproceedings{A6,
  title={Swin transformer: Hierarchical vision transformer using shifted windows},
  author={Liu, Ze and Lin, Yutong and Cao, Yue and Hu, Han and Wei, Yixuan and Zhang, Zheng and Lin, Stephen and Guo, Baining},
  booktitle={Proceedings of the IEEE/CVF international conference on computer vision},
  pages={10012--10022},
  year={2021}
}

@article{A7,
  title={An image is worth 16x16 words: Transformers for image recognition at scale},
  author={Dosovitskiy, Alexey},
  journal={arXiv preprint arXiv:2010.11929},
  year={2020}
}

@inproceedings{I1,
  title={Segment anything},
  author={Kirillov, Alexander and Mintun, Eric and Ravi, Nikhila and Mao, Hanzi and Rolland, Chloe and Gustafson, Laura and Xiao, Tete and Whitehead, Spencer and Berg, Alexander C and Lo, Wan-Yen and others},
  booktitle={Proceedings of the IEEE/CVF international conference on computer vision},
  pages={4015--4026},
  year={2023}
}

@inproceedings{I2,
  title={Depth anything: Unleashing the power of large-scale unlabeled data},
  author={Yang, Lihe and Kang, Bingyi and Huang, Zilong and Xu, Xiaogang and Feng, Jiashi and Zhao, Hengshuang},
  booktitle={Proceedings of the IEEE/CVF Conference on Computer Vision and Pattern Recognition},
  pages={10371--10381},
  year={2024}
}

@article{I3,
  title={Depth anything v2},
  author={Yang, Lihe and Kang, Bingyi and Huang, Zilong and Zhao, Zhen and Xu, Xiaogang and Feng, Jiashi and Zhao, Hengshuang},
  journal={Advances in Neural Information Processing Systems},
  volume={37},
  pages={21875--21911},
  year={2025}
}

@article{I4,
  title={Language models are few-shot learners},
  author={Brown, Tom and Mann, Benjamin and Ryder, Nick and Subbiah, Melanie and Kaplan, Jared D and Dhariwal, Prafulla and Neelakantan, Arvind and Shyam, Pranav and Sastry, Girish and Askell, Amanda and others},
  journal={Advances in neural information processing systems},
  volume={33},
  pages={1877--1901},
  year={2020}
}

@inproceedings{I5,
  title={Label-free Anomaly Detection in Aerial Agricultural Images with Masked Image Modeling},
  author={Shikhar, Sambal and Sobti, Anupam},
  booktitle={Proceedings of the IEEE/CVF Conference on Computer Vision and Pattern Recognition},
  pages={5440--5449},
  year={2024}
}

@article{I6,
  title={Masking augmentation for supervised learning},
  author={Heo, Byeongho and Kim, Taekyung and Yun, Sangdoo and Han, Dongyoon},
  journal={arXiv preprint arXiv:2306.11339},
  year={2023}
}

@article{I7,
  title={Evolved Hierarchical Masking for Self-Supervised Learning},
  author={Feng, Zhanzhou and Zhang, Shiliang},
  journal={IEEE Transactions on Pattern Analysis and Machine Intelligence},
  year={2024},
  publisher={IEEE}
}

@inproceedings{I8,
  title={Mergevq: A unified framework for visual generation and representation with disentangled token merging and quantization},
  author={Li, Siyuan and Zhang, Luyuan and Wang, Zedong and Tian, Juanxi and Tan, Cheng and Liu, Zicheng and Yu, Chang and Xie, Qingsong and Lu, Haonan and Wang, Haoqian and others},
  booktitle={Proceedings of the Computer Vision and Pattern Recognition Conference},
  pages={19713--19723},
  year={2025}
}

@inproceedings{I9,
title={Pre-training with Random Orthogonal Projection Image Modeling},
author={Maryam Haghighat and Peyman Moghadam and Shaheer Mohamed and Piotr Koniusz},
booktitle={The Twelfth International Conference on Learning Representations},
year={2024},
url={https://openreview.net/forum?id=z4Hcegjzph}
}

@inproceedings{I10,
  title={MIMTrack: In-Context Tracking via Masked Image Modeling},
  author={Wang, Xingmei and Nie, Guohao and Meng, Jiaxiang and Yan, Zining},
  booktitle={Proceedings of the AAAI Conference on Artificial Intelligence},
  volume={39},
  number={8},
  pages={7979--7987},
  year={2025}
}

@article{I11,
  title={Masked image modeling: A survey},
  author={Hondru, Vlad and Croitoru, Florinel Alin and Minaee, Shervin and Ionescu, Radu Tudor and Sebe, Nicu},
  journal={International Journal of Computer Vision},
  pages={1--47},
  year={2025},
  publisher={Springer}
}

@inproceedings{T1,
  title={Detrs beat yolos on real-time object detection},
  author={Zhao, Yian and Lv, Wenyu and Xu, Shangliang and Wei, Jinman and Wang, Guanzhong and Dang, Qingqing and Liu, Yi and Chen, Jie},
  booktitle={Proceedings of the IEEE/CVF conference on computer vision and pattern recognition},
  pages={16965--16974},
  year={2024}
}

@software{T2,
author = {Jocher, Glenn and Qiu, Jing and Chaurasia, Ayush},
license = {AGPL-3.0},
month = jan,
title = {{Ultralytics YOLO}},
url = {https://github.com/ultralytics/ultralytics},
version = {8.0.0},
year = {2023}
}

@article{T3,
  title={The pascal visual object classes (voc) challenge},
  author={Everingham, Mark and Van Gool, Luc and Williams, Christopher KI and Winn, John and Zisserman, Andrew},
  journal={International journal of computer vision},
  volume={88},
  pages={303--338},
  year={2010},
  publisher={Springer}
}

@inproceedings{T4,
  title={Microsoft coco: Common objects in context},
  author={Lin, Tsung-Yi and Maire, Michael and Belongie, Serge and Hays, James and Perona, Pietro and Ramanan, Deva and Doll{\'a}r, Piotr and Zitnick, C Lawrence},
  booktitle={Computer vision--ECCV 2014: 13th European conference, zurich, Switzerland, September 6-12, 2014, proceedings, part v 13},
  pages={740--755},
  year={2014},
  organization={Springer}
}

@article{T5,
  title={Semantic understanding of scenes through the ade20k dataset},
  author={Zhou, Bolei and Zhao, Hang and Puig, Xavier and Xiao, Tete and Fidler, Sanja and Barriuso, Adela and Torralba, Antonio},
  journal={International Journal of Computer Vision},
  volume={127},
  pages={302--321},
  year={2019},
  publisher={Springer}
}

@inproceedings{T6,
  title={SegFormer: Simple and Efficient Design for Semantic Segmentation with Transformers},
  author={Xie, Enze and Wang, Wenhai and Yu, Zhiding and Anandkumar, Anima and Alvarez, Jose M and Luo, Ping},
  booktitle={Neural Information Processing Systems (NeurIPS)},
  year={2021}
}

@inproceedings{T7,
  title={Encoder-decoder with atrous separable convolution for semantic image segmentation},
  author={Chen, Liang-Chieh and Zhu, Yukun and Papandreou, George and Schroff, Florian and Adam, Hartwig},
  booktitle={Proceedings of the European conference on computer vision (ECCV)},
  pages={801--818},
  year={2018}
}

@inproceedings{Appendix1,
  title={Fast r-cnn},
  author={Girshick, Ross},
  booktitle={Proceedings of the IEEE international conference on computer vision},
  pages={1440--1448},
  year={2015}
}

@inproceedings{Appendix2,
  title={Grad-cam: Visual explanations from deep networks via gradient-based localization},
  author={Selvaraju, Ramprasaath R and Cogswell, Michael and Das, Abhishek and Vedantam, Ramakrishna and Parikh, Devi and Batra, Dhruv},
  booktitle={Proceedings of the IEEE international conference on computer vision},
  pages={618--626},
  year={2017}
}

@article{F1,
  title={Learning deep representations by mutual information estimation and maximization},
  author={Hjelm, R Devon and Fedorov, Alex and Lavoie-Marchildon, Samuel and Grewal, Karan and Bachman, Phil and Trischler, Adam and Bengio, Yoshua},
  journal={arXiv preprint arXiv:1808.06670},
  year={2018}
}

@inproceedings{F2,
  title={The unreasonable effectiveness of deep features as a perceptual metric},
  author={Zhang, Richard and Isola, Phillip and Efros, Alexei A and Shechtman, Eli and Wang, Oliver},
  booktitle={Proceedings of the IEEE conference on computer vision and pattern recognition},
  pages={586--595},
  year={2018}
}

@article{T8,
  title={Infrared and visible image fusion via saliency analysis and local edge-preserving multi-scale decomposition},
  author={Zhang, Xiaoye and Ma, Yong and Fan, Fan and Zhang, Ying and Huang, Jun},
  journal={Journal of the Optical Society of America A},
  volume={34},
  number={8},
  pages={1400--1410},
  year={2017},
  publisher={Optical Society of America}
}

@article{T9,
  title={Perceptual fusion of infrared and visible images through a hybrid multi-scale decomposition with Gaussian and bilateral filters},
  author={Zhou, Zhiqiang and Wang, Bo and Li, Sun and Dong, Mingjie},
  journal={Information fusion},
  volume={30},
  pages={15--26},
  year={2016},
  publisher={Elsevier}
}

@inproceedings{T10,
  title={Crossvit: Cross-attention multi-scale vision transformer for image classification},
  author={Chen, Chun-Fu Richard and Fan, Quanfu and Panda, Rameswar},
  booktitle={Proceedings of the IEEE/CVF international conference on computer vision},
  pages={357--366},
  year={2021}
}

@inproceedings{T11,
  title={Selective refinement network for high performance face detection},
  author={Chi, Cheng and Zhang, Shifeng and Xing, Junliang and Lei, Zhen and Li, Stan Z and Zou, Xudong},
  booktitle={Proceedings of the AAAI conference on artificial intelligence},
  volume={33},
  number={01},
  pages={8231--8238},
  year={2019}
}

@inproceedings{T12,
  title={Early versus late fusion in semantic video analysis},
  author={Snoek, Cees GM and Worring, Marcel and Smeulders, Arnold WM},
  booktitle={Proceedings of the 13th annual ACM international conference on Multimedia},
  pages={399--402},
  year={2005}
}

@article{T13,
  title={Multimodal machine learning: A survey and taxonomy},
  author={Baltru{\v{s}}aitis, Tadas and Ahuja, Chaitanya and Morency, Louis-Philippe},
  journal={IEEE transactions on pattern analysis and machine intelligence},
  volume={41},
  number={2},
  pages={423--443},
  year={2018},
  publisher={IEEE}
}

\end{document}